\documentclass[11pt,onecolumn]{article}

\usepackage{graphicx}
\usepackage{amsmath, amssymb, amsthm}
\usepackage[utf8]{inputenc}
\usepackage[left=1in, right=1in, top=1in, bottom=1in]{geometry}
\usepackage{natbib}
\usepackage{booktabs}
\usepackage{subcaption}
\usepackage{algorithm}
\usepackage{algpseudocode}
\usepackage[hidelinks]{hyperref}
\usepackage{xcolor}
\usepackage{tikz}
\usepackage{stmaryrd}
\usepackage{tabularx}
\usepackage{setspace}
\usepackage{authblk}

\newtheorem{theorem}{Theorem}

\newcommand{\tprM}{\mathrm{TPR}}
\newcommand{\fprM}{\mathrm{FPR}}

\setcitestyle{numbers,square}

\usetikzlibrary{arrows.meta, calc, positioning}

\title{Simulating Classification Models for Ex-Ante Evaluation of Predict-Then-Optimize Methods}
\author{Pieter Smet\thanks{Corresponding author, \url{pieter.smet@kuleuven.be}}}
\affil{KU Leuven, Department of Computer Science, NUMA \\ Gebroeders De Smetstraat 1, 9000 Gent, Belgium}
\date{}

\begin{document}

\maketitle

\begin{abstract}
    Predict-Then-Optimize combines machine learning predictions with downstream optimization to support decision-making when problem parameters are unknown at the time of solving.
    However, better predictive performance does not necessarily lead to better decisions, making it useful to assess this relationship before investing in the development of a prediction model.
    Existing simulation-based approaches enable such ex-ante evaluation, but are limited to binary classification and may require solving the downstream optimization problem many times.
    We generalize this methodology to optimization problems with categorical uncertain parameters by introducing a method for simulating multiclass predictions at prescribed performance levels and using it to construct a prediction-error-to-decision-regret mapping.
    To reduce the computational effort required to obtain this mapping, we also propose a first-order approximation based on the regret caused by individual misclassifications. 
    Computational experiments confirm that the proposed prediction simulation algorithm reproduces the target classification performance and that the first-order approximation closely matches the simulation-based error-to-regret mapping for some problems.
    Its accuracy decreases when interactions between simultaneous misclassifications become more important.
    These results demonstrate the potential of the proposed approach and identify new questions about when simple approximations of the error-to-regret relationship are sufficiently accurate.
\end{abstract}

\section{Introduction}

In many practical decision support systems, decisions are obtained by solving combinatorial optimization problems such as, for example, routing, scheduling or resource allocation.
A common strategy to manage uncertainty in such systems is to combine machine learning (ML) with combinatorial optimization techniques.
An established example of this approach is Predict-Then-Optimize (PTO) \cite{sadana2025survey}.
To handle parameters of the optimization model that are unknown at the time a decision must be made, PTO first uses a prediction model to make point predictions for these parameters.
These predictions are then included in the optimization model, which is subsequently solved to arrive at decisions that optimize a given objective function.

In traditional PTO, the prediction model is trained independently from the optimization model, with the aim of maximizing predictive accuracy.
This approach implicitly assumes that more accurate predictions lead to better downstream decisions.
However, several studies have shown that this is not necessarily always the case, thereby arguing that training prediction models using alternative loss functions can be beneficial for the downstream optimization task \cite{mandi2024decision}.
In this paper, we focus on this underlying assumption of traditional PTO and propose a methodology to verify whether it holds for constrained combinatorial optimization problems.

Our methodology centers around \textit{simulating predictions}, which we define as the use of a stochastic mechanism to generate artificial predictions that mimic the behavior of an actual ML model, without requiring the model itself to be trained.
Algorithms for simulating predictions can be parametrized to emulate prediction models at controlled levels of predictive quality.
By systematically varying these levels, generating simulated predictions and solving the corresponding downstream optimization problem, we obtain an empirical mapping of prediction quality to regret of the downstream optimization problem.
In practice, such a mapping is useful to identify the minimum performance requirements an ML model must meet to ensure that the optimized solutions reach a desired level of quality.
By relying on simulation rather than repeatedly training and evaluating actual ML models, this approach can also reduce the computational and financial costs associated with model development and assessment.

Our work builds upon the simulation-based evaluation method introduced by \citet{doneda2026robust}, who studied a PTO for nurse rostering in which the number of on-call duties was predicted using a binary classifier.
We generalize their algorithm for simulating binary classifiers by introducing a new method for multiclass classification and establish consistency and concentration results of this new algorithm.
In a computational study, we confirm the theoretical properties of our new algorithm.
We also formalize the simulation-based evaluation method of \citet{doneda2026robust} and introduce a computationally less demanding alternative based on first-order approximation of the prediction-error-to-regret mapping.
We evaluate both the simulation-based and first-order approximation methods on three combinatorial optimization problems.
The computational results demonstrate how the first-order method can closely approximate the mapping obtained by the simulation-based method for some problems, while it fails to capture more complex interactions between misclassifications that occur in other problems.

The remainder of this paper is organized as follows.
Section \ref{sec:related_work} reviews the related literature.
Section \ref{sec:general_def} defines the concept of simulated multiclass classification, while Section \ref{sec:siml_class} introduces a new method for simulating multiclass classification predictions.
Section \ref{sec:regret-to-error} details how simulating predictions can be used to evaluate PTO methods and, in addition to the simulation-based method of \citet{doneda2026robust}, introduces a new approach based on first-order approximation.
Section \ref{sec:computational_study} presents the results of the computational study.
Finally, Section \ref{sec:conclusions} concludes the paper and outlines directions for future research.

\section{Related work}\label{sec:related_work}

Research on PTO has considered both the development of new methods and their application in a wide variety of settings. 
In this section, we focus specifically on studies concerned with evaluating the performance of PTO methods, organizing the literature into three complementary perspectives.
First, Section \ref{sec:_related_work_2} discusses theoretical results concerning the performance of PTO methods.
Second, Section \ref{sec:_related_work_1} reviews contributions that empirically evaluate PTO methods in specific use cases.
Finally, Section \ref{sec:_related_work_3} considers contributions most closely related to ours, namely approaches that enable ex-ante evaluation of PTO methods.

\subsection{Theoretical results}\label{sec:_related_work_2}


A stream of theoretical work relevant to our work has focused on characterizing the conditions under which predictive performance and downstream decision quality are not aligned.
A central observation is that a prediction error should not necessarily be evaluated independently of the optimization problem in which the prediction is used. 
\citet{elmachtoub2022smart} formalize this idea through the Smart Predict-then-Optimize (SPO) framework. 
They introduce the SPO loss, which measures the downstream decision loss induced by a prediction rather than the prediction error itself, and derive the convex SPO+ surrogate for training prediction models. 
Their results provide a theoretical foundation for decision-focused learning (DFL): prediction errors that appear similar under a conventional loss function can have substantially different consequences for the downstream optimization problem.

\citet{cameron2022perils} study when the separation of prediction and optimization can lead to suboptimal decisions. 
More specifically, they demonstrate that the suitability of PTO depends not only on predictive performance, but also on how the predicted quantities interact with each other and how they appear in the downstream optimization problem.
They consider a setting in which the prediction model is optimal for the prediction task, thereby abstracting away from estimation error and isolating the effect of the quantity being predicted. 
They show that when the objective coefficients depend linearly on the prediction targets, optimizing based on their conditional expectations is optimal. 
However, when multiple correlated prediction targets are combined nonlinearly to determine the coefficients of the optimization objective, PTO can be suboptimal even when the prediction stage is optimal. 
For certain nonlinear parameterizations, they construct instances for which the performance gap between PTO and end-to-end approaches becomes unbounded. 

A complementary perspective is provided by \citet{elmachtoub2023estimate}, who compare PTO with an integrated estimation-optimization approach. 
Their results show that integrating estimation and optimization is not always better. 
When the prediction model can represent the true relationship in the data, PTO asymptotically achieves lower regret than the integrated approach. 
However, when the model cannot fully capture the true relationship, integrating estimation and optimization can be beneficial.
This shows that the value of separating prediction from optimization does not only depend on the downstream optimization problem, but also on how well the prediction model captures the underlying uncertainty.

These theoretical results identify structural and statistical conditions under which prediction and decision quality may align or diverge. 
However, they do not quantify how changes in prediction quality translate into downstream decision regret.
The empirical evaluation approaches reviewed in the following section address this question from a different perspective.


\subsection{Empirical evaluation}\label{sec:_related_work_1}

There exists a large body of work that empirically evaluates PTO methods in specific application settings. 
These studies differ in which aspect of the PTO pipeline they evaluate and how they assess its performance.
We distinguish the following five, non-mutually exclusive, types of evaluation.
An \textit{operational benchmark} (OB) evaluates the complete PTO approach by comparing the resulting decisions or operational outcomes with those obtained using alternative methods or existing practice. 
In this type of evaluation, the prediction model itself is not the primary subject of interest.
In a \textit{predictive and operational evaluation} (POE), the two stages of PTO are evaluated separately: the predictive model is assessed using conventional prediction metrics, after which the decisions obtained from its predictions are evaluated using operational or optimization-based performance measures.
A \textit{prediction-decision relationship} (PDR) evaluation goes one step further by explicitly investigating how differences in predictive performance translate into differences in downstream decision quality. 
Typically, prediction models with different levels of performance are compared using both predictive and operational metrics. 
A related evaluation is \textit{prediction-error sensitivity} (PES) analysis, in which prediction errors are systematically varied or perturbed to assess how sensitive the resulting decisions are to the magnitude or characteristics of these errors.
Finally, a \textit{method comparison} (MC) evaluates alternative approaches for integrating prediction and optimization based primarily on their downstream performance. 
This includes comparisons between different PTO formulations as well as comparisons with methods that more directly account for the downstream optimization problem.

Table \ref{tab:literature_overview} provides an overview of representative studies and classifies them based on their type of evaluation.
\begin{table}[htbp]
  \centering
  \resizebox{\textwidth}{!}{  
    \begin{tabular}{llllllll}
    \toprule
          &       &       & \multicolumn{5}{c}{Evaluation} \\
    \cmidrule{4-8}
    Reference & Predicted parameter(s) & Optimization problem & OB & PDR & PES & POE & MC \\
    \midrule
    \citet{smirnov2020analytics}  &  Customer arrivals, service time  &  Staffing  &       &       &       & \checkmark &  \\
    \citet{birolini2023day}  &  Flight delays  &  Aircraft routing  & \checkmark &       &       & \checkmark &  \\
    \citet{alnaggar2025distributionally}   &  Patient demand  &  Capacity expansion  & \checkmark &       &       & \checkmark &  \\
    \citet{belhaiza2026data}   &  Travel times, demand  &  Dynamic vehicle routing  & \checkmark &       &       & \checkmark &  \\
    \citet{harris2021selecting}  &  Patient no-shows  &  Appointment scheduling  &       & \checkmark &       & \checkmark &  \\
    \citet{cakir2023rollout}  &  Fish catch  &  Trawler routing  &       & \checkmark &       & \checkmark &  \\
    \citet{jacquillat2022predictive}  &  Passenger itineraries  &  Ground-delay allocation  & \checkmark &       &       &       &  \\
    \citet{van2022dynamic}  &  Machine failures  &  Maintenance scheduling  & \checkmark &       &       &       &  \\
    \citet{cao2025real}  &  Passenger demand  &  Transit scheduling  & \checkmark &       &       &       &  \\
    \citet{abdollahi2023demand}  &  Delivery orders  &  Last-mile routing  & \checkmark &       & \checkmark &       &  \\
    \citet{hu2025prediction}  &  Patient arrivals  &  Nurse staffing  &       &       & \checkmark &       &  \\
    \citet{vanderschueren2024new}  &  Task successes  &  Resource allocation  &       &       &       &       & \checkmark \\
    \citet{yan2024classification}   &  Ship deficiencies  &  Inspection allocation  &       &       &       &       & \checkmark \\
    \citet{jia2025scenario}  &  Retailer demand  &  Inventory routing  &       &       &       &       & \checkmark \\
    \bottomrule
    \end{tabular}%
  }
  \caption{Overview of empirical studies of PTO methods. OB: operational benchmark, POE: predictive and operational evaluation, PDR: prediction-decision relationship, PES: prediction error sensitivity, MC: method comparison.}
  \label{tab:literature_overview}
\end{table}%

Operational benchmarking is common in the empirical PTO literature. 
For example, \citet{jacquillat2022predictive}, \citet{van2022dynamic} and \citet{cao2025real} primarily assess whether the decisions produced by their PTO approaches improve relevant operational benchmarks. 
Several other studies combine this analysis with a separate evaluation of the predictive component. 
For example, \citet{birolini2023day}, \citet{alnaggar2025distributionally} and \citet{belhaiza2026data} evaluate both predictive performance and the resulting operational outcomes. 
\citet{smirnov2020analytics} similarly assess both components of the PTO pipeline, although without explicitly studying how changes in prediction quality translate into changes in decision quality.

Particularly relevant to our work are studies that explicitly investigate the relationship between predictive and downstream performance. 
\citet{harris2021selecting} and \citet{cakir2023rollout} evaluate multiple predictive approaches using both prediction and decision metrics, thereby examining whether improvements in predictive performance result in improvements in the subsequent decisions. 
Both studies illustrate why evaluating a prediction model only using conventional predictive metrics may be insufficient in a PTO setting. 
While predictive and operational performance can be evaluated separately, their rankings do not necessarily match.

A second way of studying this relationship is through prediction error sensitivity analysis. 
Rather than only comparing the errors produced by a fixed set of trained prediction models, these studies deliberately vary characteristics of the prediction errors and examine the consequences for the downstream problem. 
\citet{abdollahi2023demand} perform such an analysis in the context of last-mile routing, while \citet{hu2025prediction} study the effect of prediction errors on nurse staffing decisions. 

Other studies focus primarily on comparing alternative ways of combining prediction and optimization. 
\citet{vanderschueren2024new}, \citet{yan2024classification} and \citet{jia2025scenario} compare different predictive-prescriptive approaches based on the quality of the resulting decisions. 
Such comparisons are closely related to the literature on DFL, in which the predictive model is trained while taking the downstream optimization objective into account, rather than being trained independently using only a conventional prediction loss \citep{mandi2024decision}.

Most of the studies reported in Table \ref{tab:literature_overview} consider a particular application, dataset, prediction approach and downstream optimization problem. 
Their conclusions therefore primarily characterize the PTO setting under consideration. 
\citet{geng2024benchmarking} take an important step toward more systematic empirical evaluation by proposing a benchmarking framework that compares several PTO and related methods across eight optimization problems. 

\subsection{Ex-ante evaluation}\label{sec:_related_work_3}

The approaches discussed in Section \ref{sec:_related_work_1} evaluate the effect of predictions after a particular prediction model has been specified. 
A complementary perspective is to evaluate the potential value of predictive information before developing such a model. 
We refer to this as \emph{ex-ante evaluation}: rather than training and evaluating specific prediction models, predictions are generated according to prescribed characteristics of predictive performance, after which their consequences for the downstream decision problem are evaluated. 
By varying these characteristics, ex-ante evaluation can provide insight into whether developing a prediction model is worthwhile and what level of predictive performance would be required for it to provide sufficient operational value.

The general idea of evaluating predictive information by simulating prediction errors has been explored in the forecasting literature. 
\citet{sanders2009quantifying} study the impact of forecasting errors on organizational costs in a labor-intensive warehouse by varying both forecast bias and variability. 
\citet{fildes2011incorporating}  develop a simulation framework to separately represent demand uncertainty and forecasting error and investigate their effects on unit costs and customer service levels in supply chain planning. 
Similarly, \citet{altendorfer2016effects} examine how forecasting errors affect production planning decisions and the resulting costs. 
Although these studies do not consider PTO as defined in this paper, they establish the broader principle that the operational consequences of prediction errors can be investigated by simulating errors with controlled characteristics rather than by training and comparing specific prediction models.

\citet{farrington2024many} apply a similar idea to binary classification. 
They consider an ML-guided policy for issuing platelets in a hospital blood bank, where a classifier predicts whether issued platelets will be returned unused. 
Before developing the prediction model, they simulate hypothetical classifiers with different combinations of sensitivity and specificity and evaluate their effect on operational performance using a simulation of the blood-bank workflow. 
This simulation-based analysis allows them to characterize how classifier performance translates into operational outcomes and to identify regions of predictive performance for which the ML-guided policy improves upon the conventional issuing policy.

\citet{doneda2026robust} discuss this simulation-based idea in the context of PTO. 
They consider a personnel rostering problem in which binary predictions of employee absenteeism are used to determine the number of on-call shifts that should be included in a roster. 
Rather than training absenteeism prediction models with different levels of performance, they simulate predictions at prescribed performance levels. 
The resulting predictions are subsequently used as inputs to the rostering optimization problem, and the generated rosters are evaluated under realized employee absences. 
By systematically varying the predictive performance of the simulated classifier, they determine the performance requirements under which the PTO approach achieves a desired level of robustness and outperforms non-data-driven rostering policies. 

Our work builds directly on this simulation-based approach but extends it in several directions. 
First, whereas \citet{doneda2026robust} consider binary classification, we introduce a more general approach for simulating multiclass classification predictions at prescribed class-wise performance levels and establish theoretical guarantees on the resulting simulated performance. 
Second, we formalize the ex-ante PTO evaluation through an error-to-regret mapping that directly relates prediction quality to the expected regret of the downstream optimization problem. 
Finally, since obtaining this mapping using the methodology proposed by \citet{doneda2026robust} requires repeatedly solving the downstream optimization problem, we introduce a computationally less demanding approximation based on the effects of individual misclassifications.


\section{Definition of simulated classification}\label{sec:general_def}

Let $\mathcal{Y} = \{ 1,...,K \}$ be a finite set of labels and let $Y \in \mathcal{Y}$ denote the true class label of an instance, and $\hat{Y} \in \mathcal{Y}$ the corresponding predicted label.
A simulated classification model is a stochastic prediction generator defined by a conditional probability kernel $Q : \mathcal{Y} \times \mathcal{Y} \rightarrow [0,1]$ where 
\[
Q(\hat{y} \mid y) = \mathbb{P}(\hat{Y} = \hat{y} \mid Y = y)
\]
is the probability that the simulated model predicts label $\hat{y}$ when the true label is $y$.

Given a dataset with true labels $y_1,...,y_n$, the simulated model generates predictions $\hat{Y}_1,...,\hat{Y}_n$ by sampling independently according to
\[
    \hat{Y}_r \sim Q(\cdot \mid y_r), \qquad r = 1,...,n.
\]
In other words, a simulated classification model produces predictions by sampling from a prescribed conditional distribution over predicted labels, given the true label. 
$Q$ can be seen as a conditional probability matrix \eqref{eq:kernel}, where row $i$ contains the conditional distribution of predicted labels given the true class $i$.
The diagonal entries $Q(y \mid y)$ represent class-wise correct classification probabilities, while the off-diagonal entries $Q(\hat{y} \mid y)$ where $\hat{y} \neq y$ represent specific misclassification probabilities.

\begin{equation}
\label{eq:kernel}
Q =
\begin{pmatrix}
Q(1 \mid 1) & Q(2 \mid 1) & \cdots & Q(K \mid 1) \\
Q(1 \mid 2) & Q(2 \mid 2) & \cdots & Q(K \mid 2) \\
\vdots & \vdots & \ddots & \vdots \\
Q(1 \mid K) & Q(2 \mid K) & \cdots & Q(K \mid K)
\end{pmatrix}
\end{equation}

We use the class-wise true positive rate (TPR) and one-vs-rest false positive rate (FPR) to characterize the performance of the classification models we want to simulate.
For each class $i \in \mathcal{Y}$, $\tprM_i$ is the probability that the model predicts class $i$ when the true class is also $i$, denoted as $\mathbb{P}(\hat{Y} = i \mid Y = i)$.
For each class $j \in \mathcal{Y}$, $\fprM_j$ is the probability that the model predicts class $j$ when the true class is not $j$, denoted as $\mathbb{P}(\hat{Y} = j \mid Y \neq j)$.

Let $\alpha_i = \tprM_i$ and $\beta_j = \fprM_j$ denote the target true positive rate and one-vs-rest false positive rate, respectively.
The kernel $Q$ must be constructed so that the following constraints are satisfied:
\begin{itemize}
    \item Each row of $Q$ must be a valid probability distribution:
    \[
        Q(j \mid i) \geq 0 \qquad \forall i,j \in \mathcal{Y}
    \]
    and
    \[
        \sum\limits_{j = 1}^K Q(j \mid i) = 1 \qquad \forall i \in \mathcal{Y}
    \]
    \item The diagonal matches the target $\mathrm{TPR}$:
    \[
        Q(i \mid i) = \alpha_i \qquad \forall i \in \mathcal{Y}
    \]
    \item The one-vs-rest $\mathrm{FPR}$ for class $j$ matches the target $\mathrm{FPR}$\footnote{The derivation of this constraint is provided in Appendix \ref{appendix:constraint3}.}:
    \[
        \sum_{\substack{i=1\\ i\neq j}}^{K} \frac{\pi_i}{1 - \pi_j} Q(j \mid i) = \beta_j \qquad \forall j \in \mathcal{Y}
    \]
    where $\pi_i = \mathbb{P}(Y = i)$ is the class proportion. In a fixed dataset with $n_i$ instances of class $i$, we use $\pi_i = n_i / n$ as the empirical proportion of class $i$.
\end{itemize}

Note that not all combinations of target values for TPR and FPR are possible.
We provide the necessary conditions for feasibility in Section \ref{sec:feasibility}.


\section{Simulating classification predictions}\label{sec:siml_class}

Building on the conditional confusion kernel $Q$ introduced in Section \ref{sec:general_def}, this section now introduces an approach for generating multiclass predictions with prescribed class-wise true positive rates and one-vs-rest false positive rates. 
Section \ref{sec:full_algo} presents the construction and sampling algorithms. 
Section \ref{sec:feasibility} characterizes the feasibility of the performance targets.
Sections \ref{sec:exactness} and \ref{sec:concentration_bounds} establish exactness in expectation and finite-sample concentration of the proposed approach.

\subsection{Algorithm}\label{sec:full_algo}

We now introduce an algorithm to construct a conditional confusion kernel $Q$ satisfying the constraints listed in Section \ref{sec:general_def}.
Our algorithm does this by first constructing a joint confusion mass matrix $M$, defined as
\[
M_{ij} = \mathbb{P}(Y = i, \hat{Y} = j).
\]

Since $\mathbb{P}(Y=i, \hat{Y}=j) = \mathbb{P}(Y=i)\mathbb{P}(\hat{Y}=j \mid Y=i)$ and $\mathbb{P}(\hat{Y}=j \mid Y=i) = Q(j \mid i)$, we can relate $M$ and $Q$ as
\[
M_{ij} = \pi_iQ(j \mid i).
\]
Since the diagonal entries should satisfy $Q(i \mid i) = \alpha_i$, we set $M_{ii} = \pi_i\alpha_i$.
In row $i$, the total probability mass is $\pi_i$. 
The off-diagonal mass in row $i$ is called the \textit{misclassification supply} $s_i$ of class $i$, and must be
\[
s_i = \sum_{\substack{j=1\\ i\neq j}}^{K} M_{ij} = \pi_i(1-\alpha_i).
\]
Now consider the FPR condition for class $j$.
By definition, $\beta_j = \mathbb{P}(\hat{Y} = j \mid Y \neq j)$.
Equivalently, $\mathbb{P}(\hat{Y} = j, Y \neq j) = \mathbb{P}(Y \neq j)\beta_j$.
Since $\mathbb{P}(Y \neq j) = 1-\pi_j$, we need the following
\[
d_j = \sum_{\substack{i=1\\ i\neq j}}^{K} M_{ij} = (1-\pi_j)\beta_j.
\]
We call this the \textit{false-positive demand} $d_j$ of predicted class $j$.

In what follows, we construct $Q(\cdot\mid i)$ only for classes with $\pi_i>0$.
For any class $i$ with $\pi_i=0$, the empirical TPR is undefined and the corresponding conditional row $Q(\cdot\mid i)$ is not used by the simulator on the fixed test set. 
The problem is to allocate the off-diagonal joint misclassification mass $M_{ij}$, such that the row sums equal the misclassification supplies $s_i$ and the column sums equal the false-positive demands $d_j$.
This problem can be solved using Linear Programming (LP), where $M_{ij}$ are the decision variables.
The LP is formulated as follows:
\begin{align}
    & \sum_{\substack{j=1\\ j\neq i}}^{K} M_{ij} = s_i  & \forall i \in \mathcal{Y} \label{lp:c1} \\
    & \sum_{\substack{i=1\\ i\neq j}}^{K} M_{ij} = d_j  & \forall j \in \mathcal{Y} \label{lp:c2} \\
    & M_{ij} \geq 0 & \forall i,j \in \mathcal{Y} : i \neq j \label{lp:c3}
\end{align}
Since the goal is to find a feasible allocation of the off-diagonal joint misclassification mass, there is no objective function to optimize.
Constraints \eqref{lp:c1} ensure that the row sums equal the misclassification supplies.
Constraints \eqref{lp:c2} ensure that the column sums equal the false-positive demands.
Constraints \eqref{lp:c3} ensure that the allocated mass is non-negative.

Note that there may be multiple solutions for problem \eqref{lp:c1}-\eqref{lp:c3} as the target TPR and FPR values generally do not uniquely determine the conditional kernel $Q$. 
Different feasible off-diagonal allocations can yield the same TPR and FPR, but different pairwise misclassification patterns. 
If this distinction matters, the LP can be extended to include an objective function related to, for example, an application-specific misclassification preference.

The conditional kernel is constructed from $M$ as
\[
Q(j\mid i) = \frac{M_{ij}}{\pi_i} \text{\ for\ } j \neq i \quad \text{\ and\ } \quad Q(i \mid i) = \alpha_i.
\]
For an instance $r$ with true class $y_r$, the predicted class is then generated by sampling $\hat{Y}_r \sim Q(\cdot\mid y_r)$.

To distinguish kernel construction from prediction generation, we separate the approach into two algorithms.
Algorithm~\ref{alg:mc_p1} constructs a feasible conditional confusion kernel \(Q\) from the true class labels and the target TPR and FPR vectors. 
Algorithm~\ref{alg:mc_p2} uses the resulting kernel to generate simulated predictions by sampling each prediction independently as described above.

\begin{algorithm}[htpb]
    \caption{Constructing a conditional confusion kernel}
    \label{alg:mc_p1}
    \begin{algorithmic}[1]
    \Function{ConstructKernel}{$y_1,\ldots,y_n, \alpha, \beta$}
    \Require True labels $y_1,\ldots,y_n \in \mathcal{Y}=\{1,\ldots,K\}$
    \Require Target TPR vector $\alpha=(\alpha_1,\ldots,\alpha_K)$
    \Require Target FPR vector $\beta=(\beta_1,\ldots,\beta_K)$
    \Ensure Conditional confusion kernel $Q$
    
    \For{$i\in\mathcal{Y}$} \Comment{Compute class proportions and row supplies}
        \State $n_i \gets |\{r:y_r=i\}|$
        \State $\pi_i \gets n_i/n$
        \State $s_i \gets \pi_i(1-\alpha_i)$
    \EndFor
    
    \For{$j\in\mathcal{Y}$} \Comment{Compute column demands}
        \State $d_j \gets (1-\pi_j)\beta_j$
    \EndFor
    

    \State Solve $\{ \min 0 : \sum_{j\neq i}M_{ij}=s_i, \sum_{i\neq j}M_{ij}=d_j, M_{ij}\geq 0 \text{\ for\ } i \neq j \}$ \Comment{Set off-diagonal entries in $M$}
    
    \For{$i\in\mathcal{Y}$} \Comment{Set diagonal entries in $M$}
        \State $M_{ii}\gets \pi_i\alpha_i$
    \EndFor
    
    \For{$i\in\mathcal{Y}$} \Comment{Construct conditional confusion kernel $Q$}
        \For{$j\in\mathcal{Y}$}
            \State $Q(j\mid i)\gets M_{ij}/\pi_i$
        \EndFor
    \EndFor
    
    \State \Return $Q$
    \EndFunction
    \end{algorithmic}
\end{algorithm}

\begin{algorithm}[htpb]
    \caption{Sampling simulated multiclass predictions}
    \label{alg:mc_p2}
    \begin{algorithmic}[1]
    \Function{SamplePredictions}{$y_1,\ldots,y_n,Q$}
    \Require True labels $y_1,\ldots,y_n \in \mathcal{Y}=\{1,\ldots,K\}$
    \Require Conditional confusion kernel $Q(j\mid i)=\Pr(\widehat{Y}=j\mid Y=i)$
    \Ensure Simulated predictions $\hat y_1,\ldots,\hat y_n$    
    \For{$r=1,\ldots,n$}
        \State Sample $\hat y_r\sim Q(\cdot\mid y_r)$
    \EndFor    
    \State \Return $\hat y_1,\ldots,\hat y_n$
    \EndFunction
    \end{algorithmic}
\end{algorithm}

Note that in the special case of binary classification where $K=2$, the conditional confusion kernel $Q$ can be determined without computing $M$.
Instead, the target TPR and FPR values can directly be used in $Q$.
A detailed description of this, including an interpretation of the algorithm from \citet{doneda2026robust} to simulate binary classification as a conditional confusion kernel is presented in Appendix \ref{appendix:binary}.

\subsection{Feasibility conditions}
\label{sec:feasibility}

A feasible solution for problem \eqref{lp:c1}-\eqref{lp:c3} only exists if the target TPR and FPR values are compatible with the class proportions.
In other words, the target values must allow a valid joint confusion mass matrix, and hence a valid conditional confusion kernel.

First, the total misclassification supply must equal the total false-positive demand. 
Every misclassified instance creates exactly one false negative for its true class and one false positive for its predicted class.
Therefore, the total false-negative mass and the total false-positive mass must be equal:
\[
\sum\limits_{i=1}^K s_i = \sum\limits_{j=1}^Kd_j.
\]
In addition, the diagonal entries are excluded from the off-diagonal allocation. 
Since the misclassification supply of class $i$ cannot be assigned back to class $i$, it must be possible to allocate this supply to the false-positive demands of the other classes:
\[
s_i \leq \sum_{\substack{j=1\\ j\neq i}}^{K} d_j \qquad \forall i.
\]
Similarly, the false-positive demand for class $j$ cannot be supplied by instances whose true class is also $j$, so we need
\[
d_j \leq \sum_{\substack{i=1\\ i\neq j}}^{K} s_i \qquad \forall j.
\]
If these conditions are violated, no valid multiclass joint confusion matrix $M$ and therefore no conditional confusion kernel $Q$, can simultaneously have the specified class proportions, and target TPR and FPR values.

\subsection{Exactness in expectation}\label{sec:exactness}

We now show that the proposed approach for simulating predictions is exact in expectation with respect to both the class-wise true positive rates and the one-vs-rest false positive rates. 
Throughout this section, the true labels $y_1,...,y_n$ are treated as fixed, and expectations are taken only over the random predictions generated by the simulator. 
The empirical TPR for class $i$ is considered only when $n_i>0$, and the empirical one-vs-rest FPR for class $j$ is considered only when $n_j<n$.

For class $i \in \mathcal{Y}$, define the empirical simulated TPR as
\[
\widehat{\tprM}_i = \frac{1}{n_i}\sum\limits_{r:y_r=i}\mathbf{1}\{ \hat{Y}_r = i\}.
\]

\begin{theorem}[Unbiasedness of the simulated TPR]\label{th:exp_tpr}
    For any class $i$ with $n_i>0$, the predictions generated by Algorithm \ref{alg:mc_p2} satisfy
    \[
    \mathbb{E}[\widehat{\tprM}_i] = \alpha_i.
    \]
\end{theorem}
\noindent
\emph{Proof.} See Appendix~\ref{app_proof_exp_tpr}.

For class $j$, define the empirical simulated one-vs-rest FPR as
\[
\widehat{\fprM}_j = \frac{1}{n - n_j} \sum\limits_{r:y_r\neq j}\mathbf{1} \{ \hat{Y}_r = j\}.
\]

\begin{theorem}[Unbiasedness of the simulated one-vs-rest FPR]\label{th:exp_fpr}
    For any class $j$ with $n_j<n$, the predictions generated by Algorithm \ref{alg:mc_p2} satisfy
    \[
    \mathbb{E}[\widehat{\fprM}_j] = \beta_j
    \]
\end{theorem}
\noindent
\emph{Proof.} See Appendix~\ref{app_proof_exp_fpr}.


\subsection{Concentration bounds}\label{sec:concentration_bounds}

We now show that the empirical simulated TPR and FPR values concentrate around their requested target values.
As in Section \ref{sec:exactness}, the true labels $y_1,...,y_n$ are treated as fixed, and probabilities are taken only over the random predictions generated by Algorithm \ref{alg:mc_p2}. 
The empirical TPR bound is stated only for classes with $n_i > 0$, and the empirical one-vs-rest FPR bound is stated only for classes with $n_j < n$.

\begin{theorem}[Concentration of the simulated TPR]\label{th:conc_tpr}
    For any class $i \in \mathcal{Y}$ with $n_i>0$ and any $\epsilon > 0$, the predictions generated by Algorithm \ref{alg:mc_p2} satisfy
    \[
    \mathbb{P}\left( |\widehat{\tprM}_i - \alpha_i| \geq \epsilon \right) \leq 2 \exp(-2n_i\epsilon^2)
    \]
\end{theorem}
\noindent
\emph{Proof.} See Appendix~\ref{app_proof_conc_tpr}.

\begin{theorem}[Concentration of the simulated FPR]\label{th:conc_fpr}
    For any class $j \in \mathcal{Y}$ with $n_j < n$ and any $\epsilon > 0$, the predictions generated by Algorithm \ref{alg:mc_p2} satisfy
    \[
    \mathbb{P}\left( |\widehat{\fprM}_j - \beta_j| \geq \epsilon \right) \leq 2 \exp(-2(n-n_j)\epsilon^2)
    \]
\end{theorem}
\noindent
\emph{Proof.} See Appendix~\ref{app_proof_conc_fpr}.

These bounds show that the empirical simulated TPR and FPR values concentrate exponentially fast around the requested target values.
The rate depends on the number of relevant observations, which is $n_i$ for $\widehat{\mathrm{TPR}}_i$, and $n-n_j$ for $\widehat{\mathrm{FPR}}_j$.

Class imbalance affects these bounds through the effective sample sizes.
If class $i$ is a minority class, the TPR bound may be weak because $n_i$ is small, so $\widehat{\mathrm{TPR}}_i$ can vary substantially across simulation runs. 
In contrast, for a majority class with large $n_i$, the TPR estimate concentrates much faster.
Since the exponent for the bound on $\widehat{\mathrm{FPR}}_j$ is different, minority classes usually have better FPR concentration, while majority classes may have weaker FPR concentration.
In other words, rare classes may have noisy empirical TPRs but stable empirical FPRs, while dominant classes may have stable empirical TPRs but noisier empirical FPRs.



\section{PTO error-to-regret mapping}\label{sec:regret-to-error}

We now study how the simulation approach from Section \ref{sec:siml_class} can be used to evaluate PTO methods.
Let $\xi=(\xi_1,...,\xi_n)$ denote the vector of true uncertain inputs to the downstream optimization problem, where $\xi_r \in \mathcal{Y}$ is the categorical label of instance $r$\footnote{We only consider uncertain inputs that can be represented as categorical labels.}.
Thus, $\xi$ corresponds to the vector of true labels $y$ used in Section \ref{sec:siml_class}, while $\hat{\xi}$ denotes the corresponding vector of predicted labels.
In PTO, the uncertain inputs are first predicted, after which the predicted input vector $\hat{\xi}$ is used to solve the downstream optimization problem:
\[
x^*(\hat{\xi}) \in \arg \min\limits_{x \in X} f(x, \hat{\xi}).
\]

To study the relationship between prediction quality and the quality of the resulting decision, we introduce a prediction-error-to-regret mapping that links the predictive performance of the simulated classifier, characterized by its target TPR and FPR values, to the expected regret of the downstream optimization problem. 

For a prediction vector $\hat{\xi}$, we define the regret as
\[
R(\hat{\xi},\xi) = f(x^*(\hat{\xi}),\xi) - f(x^*(\xi),\xi).
\]
The first term evaluates the decision obtained using the predicted inputs under the true inputs $\xi$, while the second term is the objective value obtained when the true inputs are known.

Let $Q_{\alpha,\beta}$ denote a feasible conditional confusion kernel constructed for the target TPR and FPR vectors $\alpha$ and $\beta$, respectively.
The expected regret induced by $Q_{\alpha,\beta}$ is then defined as
\[
\mathcal{R}(Q_{\alpha, \beta} ; \xi) = \mathbb{E}_{\hat{\xi} \sim Q_{\alpha, \beta}(\cdot \mid \xi)} \left[ R(\hat{\xi},\xi) \right] = \mathbb{E}_{\hat{\xi} \sim Q_{\alpha, \beta}(\cdot \mid \xi)} \left[ f(x^*(\hat{\xi}),\xi) - f(x^*(\xi),\xi) \right].
\]
For a fixed true input vector $\xi$ and target TPR and FPR vectors $\alpha$ and $\beta$, we define the error-to-regret mapping as
\[
\Phi(\alpha,\beta;\xi) = \mathcal{R}(Q_{\alpha,\beta};\xi).
\]

This mapping allows us to quantify how different levels of classification error translate into decision loss and, consequently, to assess how accurate a predictive model must be to achieve a desired level of decision quality.
In the remainder of this section, we present two approaches for obtaining this mapping.
Section \ref{sec:sampling} describes the approach of \citet{doneda2026robust} based on random sampling, while Section \ref{sec:delta_costs} presents a less computationally expensive approach based on an additive approximation of the expected regret.

\subsection{Random sampling approximation}\label{sec:sampling}

Algorithm \ref{alg:random_sampling} outlines the random sampling procedure.
For each replication $b=1,...,B$, we generate a simulated predicted input vector $\hat{\xi}^{(b)} = ( \hat{\xi}^{(b)}_1,\ldots,\hat{\xi}^{(b)}_n )$.
As proposed in Section \ref{sec:full_algo}, the predicted label $\hat{\xi}^{(b)}_r$ of instance $r$ is sampled independently according to $\hat{\xi}^{(b)}_r \sim Q_{\alpha,\beta}(\cdot\mid\xi_r)$.
The prediction vectors are also sampled independently across replications.

The downstream optimization problem is then solved using the sampled predicted vector $\hat{\xi}^{(b)}$ to obtain the solution $x^*(\hat{\xi}^{(b)})$.
This solution is evaluated under the true input vector $\xi$, resulting in regret
\[
R(\hat{\xi}^{(b)},\xi)
=
f(x^*(\hat{\xi}^{(b)}),\xi)
-
f(x^*(\xi),\xi).
\]
Repeating this procedure for $B$ independent replications gives the random sampling estimator
\[
\widehat{\Phi}_B(\alpha,\beta;\xi) = \frac{1}{B} \sum_{b=1}^{B} R(\hat{\xi}^{(b)},\xi).
\]
The term $f(x^*(\xi),\xi)$ is constant across replications and therefore has to be computed only once. 
The main computational cost of the estimator consists of solving the downstream optimization problem once for each sampled prediction vector. 
Consequently, computing $\widehat{\Phi}_B(\alpha,\beta;\xi)$ requires solving the optimization problem $B$ times, in addition to the single solve using the true input vector.
The random sampling approximation can therefore become computationally expensive when $B$ is large or when many combinations of target TPR and FPR values are evaluated.


\begin{algorithm}[htbp]
\caption{Random sampling approximation of $\Phi(\alpha,\beta;\xi)$}
\label{alg:random_sampling}
\begin{algorithmic}[1]
\Function{RandomSampling}{$\xi, \alpha, \beta, B$}
\Require True input vector $\xi=(\xi_1,\ldots,\xi_n)$
\Require Target TPR vector $\alpha = (\alpha_1,...,\alpha_K)$
\Require Target FPR vector $\beta = (\beta_1,...,\beta_K)$
\Require Number of replications $B$
\Ensure Random sampling estimate $\widehat{\Phi}_B(\alpha,\beta;\xi)$
\State Get $x^*(\xi) \in \arg \min_{x\in X} f(x,\xi)$ \Comment{Solve with true inputs}
\State $z^* \leftarrow f(x^*(\xi),\xi)$ \Comment{Compute the true optimal objective value}
\State $Q_{\alpha,\beta} \gets$ \Call{ConstructKernel}{$\xi_1,\ldots,\xi_n, \alpha, \beta$} (Algorithm \ref{alg:mc_p1})
\For{$b=1,\ldots,B$}
    \State $\hat{\xi}^{(b)} \gets$ \Call{SamplePredictions}{$\xi_1,\ldots,\xi_n, Q_{\alpha,\beta}$} (Algorithm \ref{alg:mc_p2})
    \State Get $x^*(\hat{\xi}^{(b)}) \in \arg \min_{x\in X} f(x,\hat{\xi}^{(b)})$ \Comment{Solve with predicted inputs}
    \State $z^{(b)} \leftarrow f(x^*(\hat{\xi}^{(b)}),\xi)$ \Comment{Evaluate solution under the true inputs}
    \State $R^{(b)} \leftarrow z^{(b)} - z^*$ \Comment{Compute the realized regret}
\EndFor
\State $\displaystyle\widehat{\Phi}_B(\alpha,\beta;\xi) \leftarrow \frac{1}{B}\sum\limits_{b=1}^{B} R^{(b)}$ \Comment{Compute average}
\State \Return $\widehat{\Phi}_B(\alpha,\beta;\xi)$
\EndFunction
\end{algorithmic}
\end{algorithm}

\subsection{First-order interaction approximation}\label{sec:delta_costs}

The random sampling estimator from Section~\ref{sec:sampling} requires solving the downstream optimization problem for every sampled prediction vector.
We now introduce an alternative approach that approximates the expected regret using the regret associated with individual misclassifications.

For each class $i\in\mathcal{Y}$, define $S_i=\{r:\xi_r=i\}$.
Consider an instance $r\in S_i$ and a possible misclassification $i\to j$, where $j\neq i$.
Let $\xi^{\,r:i\to j}$ denote the perturbed vector obtained from $\xi$ in which the label of instance $r$ is changed from $i$ to $j$, while leaving all other instances unchanged.
The regret caused by this single misclassification is
\[
\delta_{r,ij}
=
f\bigl(x^*(\xi^{\,r:i\to j}),\xi\bigr)
-
f\bigl(x^*(\xi),\xi\bigr).
\]
The average single-misclassification regret for each class pair $i,j$ with $i\neq j$ and $n_i>0$, is computed for a randomly sampled subset of instances $\hat{S}_i \subset S_i$.
More specifically, it is defined as
\[
\Delta_{ij}
=
\frac{1}{|\hat{S}_i|}
\sum_{r\in \hat{S}_i}
\delta_{r,ij}
\]
where the size of $\hat{S}_i$ is equal to the \textit{sample size} parameter $G$.

Given $Q_{\alpha,\beta}$, an instance with true class $i$ is predicted as class $j$ with probability $Q_{\alpha,\beta}(j\mid i)$.
The expected number of misclassifications of type $i\to j$ is therefore
\[
n_iQ_{\alpha,\beta}(j\mid i).
\]
From this, we define the first-order approximation of the expected regret as
\[
\widetilde{\Phi}(\alpha,\beta;\xi)
=
\sum_{i=1}^{K}
\sum_{j\neq i}
n_iQ_{\alpha,\beta}(j\mid i)
\Delta_{ij}.
\]
We can rewrite this using the joint confusion masses $M_{ij} = \pi_iQ_{\alpha,\beta}(j\mid i)$ and class proportions $\pi_i= n_i / n$, resulting in
\[
\widetilde{\Phi}(\alpha,\beta;\xi)
=
n
\sum_{i=1}^{K}
\sum_{j\neq i}
M_{ij}\Delta_{ij}.
\]
The values $\Delta_{ij}$ depend on the true input vector $\xi$ and the downstream optimization problem, but not on $\alpha$ or $\beta$.
Once these values have been computed, the first-order approximation can therefore be evaluated for different target TPR and FPR values by constructing the corresponding confusion masses and computing a weighted sum.

The first-order approximation treats the regrets from multiple simultaneous misclassifications as the sum of their individual effects.
If the downstream optimization problem has strong combinatorial interactions, this approximation may thus become inaccurate.


\section{Computational study}\label{sec:computational_study}

This section empirically evaluates the two main components of the proposed simulation-based evaluation. 
In Section \ref{sec:simul_eval}, we first test the theoretical properties of the proposed simulation approach by analyzing how closely the generated predictions reproduce the target TPR and FPR values for a finite number of samples. 
In particular, we investigate the effects of sample size, class imbalance and the distribution of misclassification probabilities across class pairs. 
In Section \ref{sec:approx_eval}, we then evaluate the error-to-regret mapping for three combinatorial optimization problems: single-machine scheduling, knapsack and shortest path. 
Using the random sampling estimator as a benchmark, we assess the accuracy of the first-order approximation.

\subsection{Evaluation of simulated predictions} \label{sec:simul_eval}

We first evaluate the empirical behavior of the proposed approach to simulate classification predictions.
We generated scenarios in which the number of samples $n$ varies from 100 to 5000, while fixing the number of classes to $K=3$ and the class proportions to $\pi = (0.33, 0.33, 0.34)$.
The target class-wise TPR values are set to $\alpha = (0.7, 0.7, 0.7)$.

To obtain compatible target FPR values, we first construct a valid conditional confusion kernel $Q(j \mid i)=P(\hat{Y}=j \mid Y=i)$.
Its diagonal entries are set to $Q(i \mid i) = \alpha_i$ so that the target class-wise TPRs are matched by construction. 
The remaining probability mass in each row $i$, $1-\alpha_i$, is distributed uniformly over the off-diagonal entries $Q(j\mid i)$, $j\neq i$. 
This ensures that each row of $Q$ is a valid probability distribution and that the off-diagonal row mass corresponds exactly to the desired misclassification probability of class $i$.
The corresponding one-vs-rest FPR for each predicted class $j$ is then obtained from the off-diagonal column mass as
\[
\beta_j = \frac{\sum_{i \neq j}\pi_i Q(j \mid i)}{1-\pi_j}.
\]
To account for the stochastic nature of Algorithm \ref{alg:mc_p2} , each experiment is repeated 5000 times.

\subsubsection{Finite sample convergence}

\begin{figure}[htbp]
    \centering
    \includegraphics[width=1\linewidth]{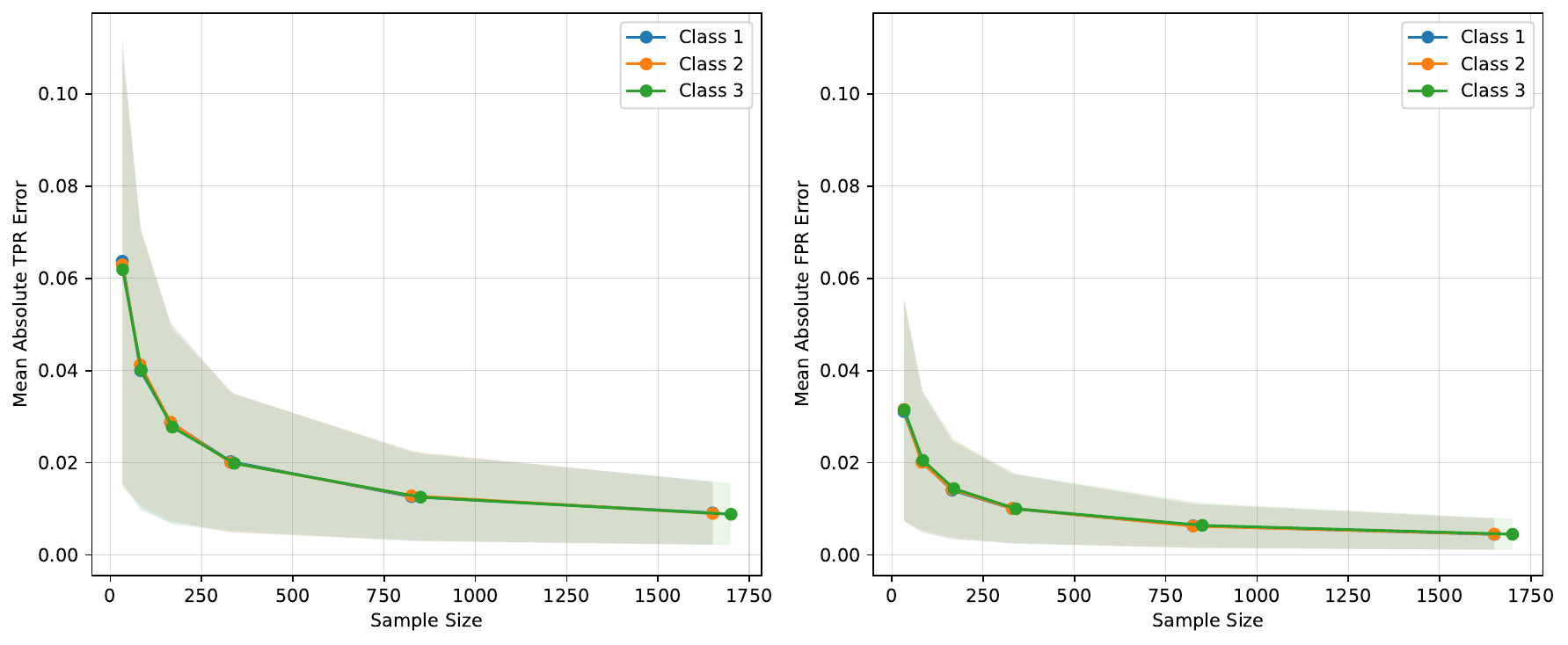}
    \caption{Finite sample convergence of TPR and FPR.}
    \label{fig:mc_results_1}
\end{figure}

We first analyze how closely the empirical TPR and FPR values produced by the simulation approach match their target values as the sample size increases.
Figure \ref{fig:mc_results_1} shows the mean absolute deviation from the target value over the 5000 repeated runs, with the shaded areas representing one standard deviation across runs.
For both TPR and FPR, the mean absolute deviation decreases as $n$ increases, following $1 / \sqrt{n}$, as expected for averages of independent Bernoulli variables. 
The lines for the three classes are nearly identical because the class proportions and target TPR values are (almost) equal.
The FPR deviations are consistently smaller than the TPR deviations because each class-wise FPR is estimated using $n-n_j$ observations, whereas the corresponding TPR is based on only $n_j$ observations.
The narrowing shaded areas demonstrate that the variability between simulation runs decreases with the sample size. 
Thus, although the predictions generated by a single run of Algorithm \ref{alg:mc_p2} may result in noticeable deviations from the target TPR and FPR values for small datasets, the generated TPR and FPR values become increasingly stable around the target values as $n$ increases.

\begin{figure}[htbp]
    \centering
    \includegraphics[width=1\linewidth]{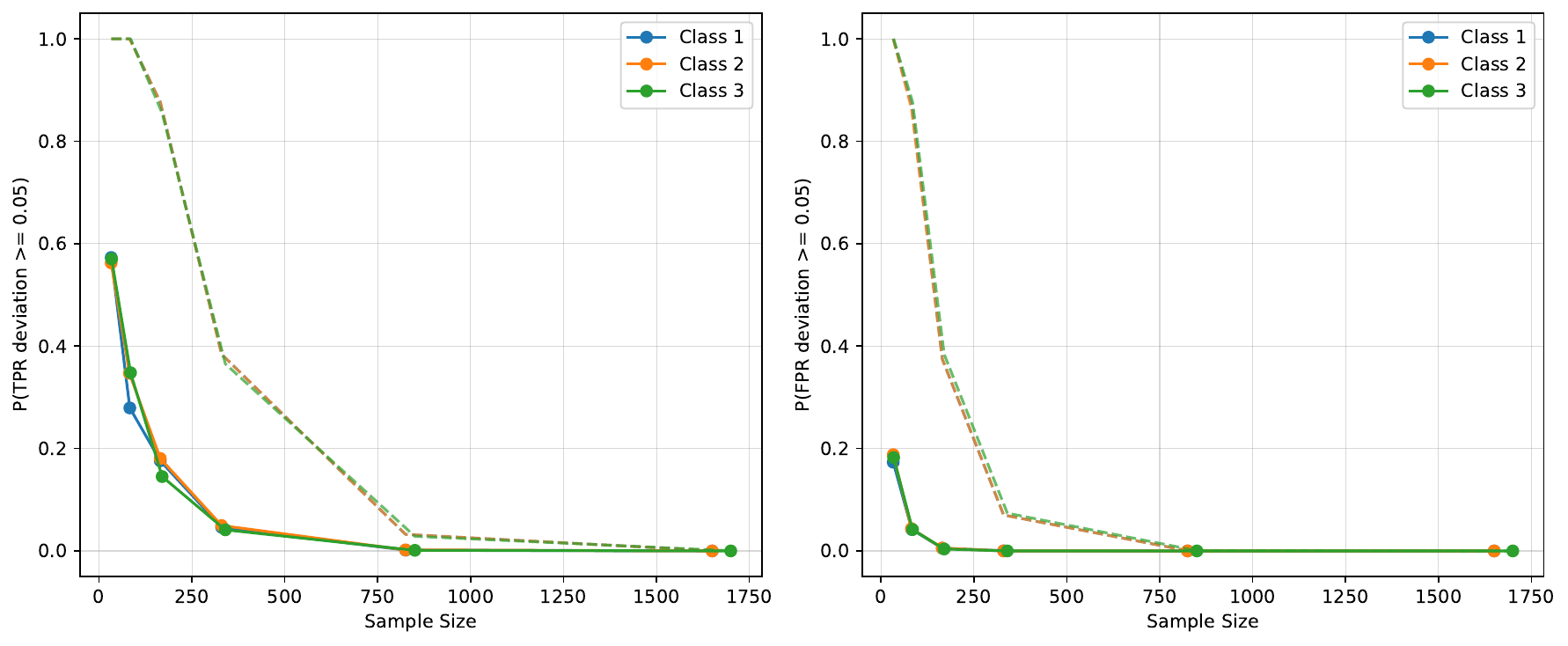}
    \caption{Concentration inequalities and bounds for TPR and FPR. Solid lines are the empirical values, dashed lines represent Hoeffding's bound.}
    \label{fig:mc_results_2}
\end{figure}

Figure \ref{fig:mc_results_2} plots the probability that the absolute deviation from the target value is at least 0.05.
For both TPR and FPR, the probabilities decrease rapidly for increasing sample size and eventually approach zero.
Similar to Figure \ref{fig:mc_results_1}, the FPR probabilities decrease more quickly because the corresponding estimates are based on larger sample sizes.
The dashed lines show the Hoeffding bounds derived in Section \ref{sec:concentration_bounds}.
The results confirm that these bounds correctly capture the decreasing probability of a substantial deviation, but they are conservative, particularly for small and intermediate sample sizes, where they are often considerably larger than the observed probabilities. 
Overall, the results support the theoretical concentration guarantees while also showing that the practical behavior of the simulation approach is considerably tighter than the bounds suggest.

\subsubsection{Effects of class imbalance}

We next examine how class imbalance affects the finite-sample variability of the simulated TPR and FPR values.
The target TPR vector remains fixed to $(0.7, 0.7, 0.7)$, while the class proportions are varied such that $\pi_1$ decreases from 0.9 to 0.05, $\pi_2$ increases from 0.05 to 0.9 and $\pi_3$ remains fixed at 0.05.
Since the proportion of the third class does not change, the analysis focuses on classes 1 and 2.

\begin{figure}[htbp]
    \centering
     \includegraphics[width=1\linewidth]{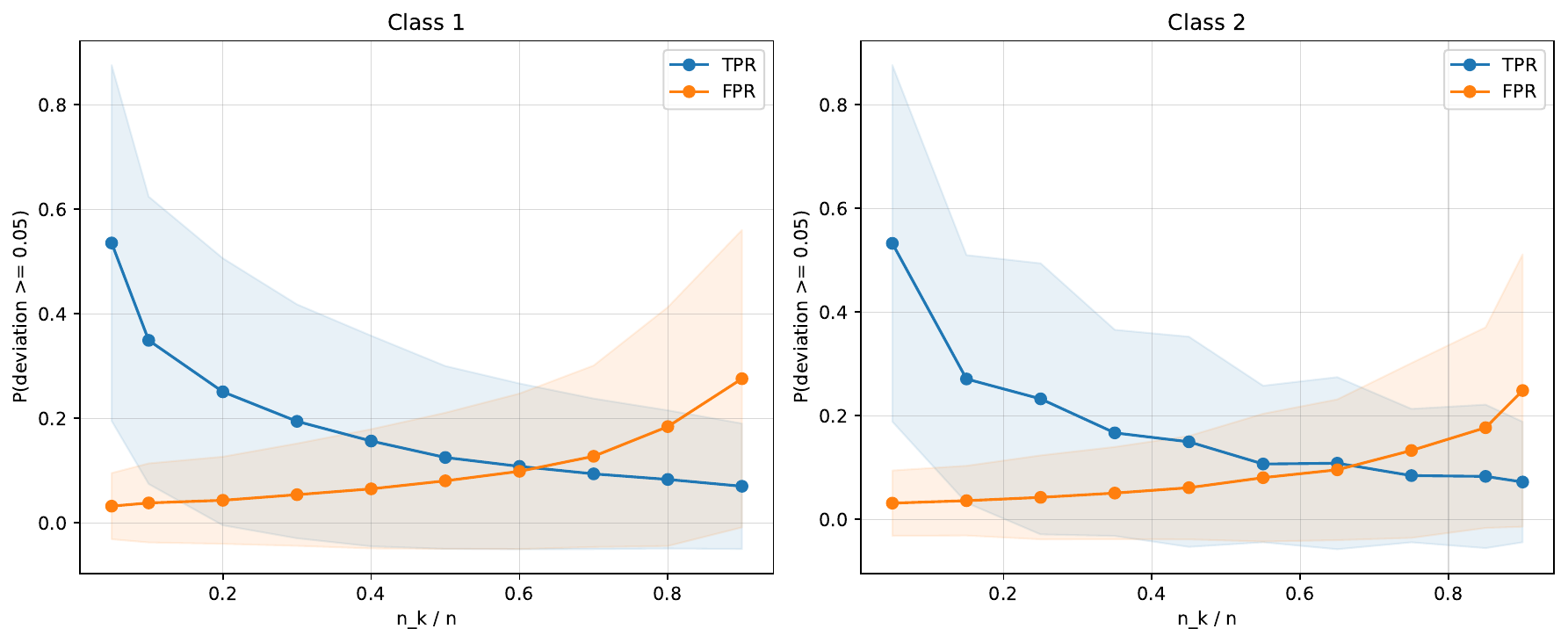}
    \caption{Impact of class imbalance.}
    \label{fig:mc_results_3}
\end{figure}

Figure \ref{fig:mc_results_3} shows, for the considered classes, the empirical probability that the absolute deviation of the simulated TPR or FPR from its target is at least 0.05, with the shaded areas representing one standard deviation across the 5000 repetitions.
As the proportion of a class increases, its empirical TPR becomes more concentrated around the target value, and the probability of a deviation of at least 0.05 decreases.
This occurs because the empirical TPR for class $k$ is estimated from $n_k$ observations, so a larger class provides a larger effective sample size. 
In contrast, the empirical one-vs-rest FPR is estimated from the $n-n_k$ observations that do not belong to class $k$. 
Its effective sample size therefore decreases as the class becomes more prevalent, causing the probability of a FPR deviation larger than 0.05 to increase.

The patterns for classes 1 and 2 are almost symmetric, reflecting the fact that their proportions vary in opposite directions. 
For rare classes, the simulated TPR shows considerable variation across runs, whereas the FPR is relatively stable. 
For dominant classes, the inverse holds: the variation for TPR is relatively small, while the FPR becomes more variable. 
These results are consistent with the concentration bounds from Section \ref{sec:concentration_bounds}, which depend on $n_k$ for TPR and $n-n_k$ for FPR.

\subsubsection{Effects of the off-diagonal misclassification structure}

Until now, we have assumed that the misclassification probability of each true class is distributed uniformly across all incorrectly predicted classes.
However, in multiclass classification, a fixed set of class-wise TPR values does not uniquely determine how prediction errors are allocated among these classes.
This distinction is important because different error allocations may lead to different empirical FPR behavior.
We therefore investigate whether the allocation of misclassification probability across class pairs affects the finite-sample behavior of the simulated FPR values.
For fixed target TPR values $\alpha=(0.7, 0.7, 0.7)$ and class proportions $\pi=(0.33, 0.33, 0.34)$, the diagonal entries of the conditional confusion kernel are fixed as $Q(i\mid i)=\alpha_i$, while the remaining probability mass $1-\alpha_i$ must be distributed over the incorrect predicted classes.

In the baseline setting, this mass is allocated uniformly, so that all incorrect labels are equally likely for a given true class:
\[
Q(j\mid i)=\frac{1-\alpha_i}{K-1},
\qquad j\neq i.
\]
However, such a uniform allocation may not reflect the errors of real multiclass classifiers, where certain pairs of classes are more easily confused than others.
We therefore also generate non-uniform off-diagonal structures.

To do so, we assume that errors are more likely between nearby classes.
For true class $i$, the relative weight assigned to an incorrect class $j$ is $w_{ij} = h^{-(|i-j|-1)}$, where $h \geq 1$ is the so-called \textit{concentration parameter}.
The weights are normalized so that the total off-diagonal probability in row $i$ equals $1-\alpha_i$.
When $h = 1$, the allocation is uniform. 
For $h > 1$, each additional step in class distance reduces the relative probability of confusion by a factor $h$. 
We vary $h \in \{1, 2, 4\}$, resulting in the FPR values shown in Table \ref{tab:frp_h}.

\begin{table}[htpb]
    \centering
    \begin{tabular}{c c c c}
    \toprule
        & \multicolumn{3}{c}{Target FPR} \\
    \cmidrule{2-4}
    $h$ & Class 1 & Class 2 & Class 3 \\
    \midrule
     1 & 0.150 & 0.150 & 0.150 \\
     2 & 0.125 & 0.200 & 0.125 \\
     4 & 0.104 & 0.240 & 0.105 \\
    \bottomrule
    \end{tabular}
    \caption{Target FPR values for different values of $h$.}
    \label{tab:frp_h}
\end{table}

\begin{figure}[htbp]
    \centering
    \includegraphics[width=1\linewidth]{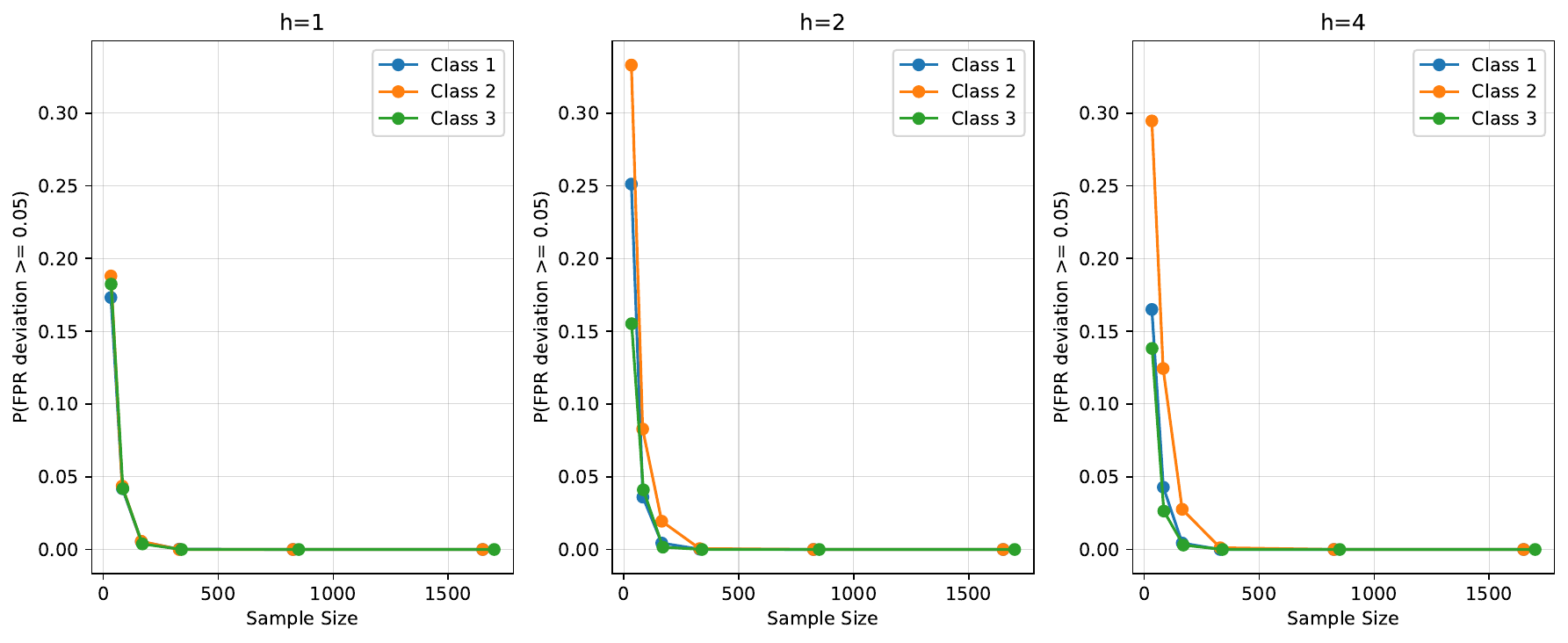}
    \caption{Impact of off-diagonal allocation.}
    \label{fig:mc_results_4}
\end{figure}

Figure \ref{fig:mc_results_4} plots the empirical probability that the absolute deviation of each class-specific simulated FPR from its target is at least 0.05, for different values of $h$.
The results show that the off-diagonal allocation affects the finite-sample variability of the class-specific FPRs, but not the consistency of the simulation algorithm. 
As $h$ increases, more false positives are shifted towards the middle class, making the empirical FPR of class 2 more variable in small samples, while those of classes 1 and 3 generally become less variable.
Nevertheless, there is no systematic increase in deviation probability as $h$ increases.
For all considered values of $h$, the deviation probabilities rapidly approach zero with increasing sample size.
Thus, the off-diagonal structure affects the finite-sample concentration of the class-specific FPRs, but not their convergence to the target values.



\subsection{Evaluation of methods to compute error-to-regret mappings}\label{sec:approx_eval}

In the following experiments, we evaluate the methods from Section \ref{sec:regret-to-error} for obtaining the error-to-regret mapping.
The analysis considers three combinatorial optimization problems: single-machine scheduling, one-dimensional knapsack, and shortest path.
Table \ref{tab:cases} provides an overview of the considered problems and the categorical uncertain input data that is predicted before the optimization problem is solved.
These problems differ in type of decisions, constraints and objective function, providing a diverse test bed for assessing the different approaches.
For each problem, we examine how changes in predictive performance translate into downstream regret. 
We also evaluate the accuracy of the first-order interaction approximation using the random sampling estimator as a benchmark.

\begin{table}[htbp]
    \centering
    \begin{tabular}{l l}
    \toprule
     Problem & Unknown input data \\
    \midrule
    Single machine scheduling & Job priority that determines completion-time weight \\
    One-dimensional knapsack & Item type that determines item profit \\
    Shortest path on road networks & Road class that determines increase in travel time\\
    \bottomrule
    \end{tabular}
    \caption{Overview of the considered optimization problems.}
    \label{tab:cases}
\end{table}

The experimental procedure is the same for each problem. 
First, we generate a set of problem instances by varying characteristics that are expected to impact both the error-to-regret mapping and the accuracy of the first-order approximation. 
Then, for each instance, we vary the target TPR values, construct compatible target FPR values, and estimate the resulting expected regret using the methods from Section \ref{sec:regret-to-error}. 

\subsubsection{Single-machine scheduling with job priorities} 

We first consider a single-machine scheduling problem in which the objective is to minimize the total weighted completion time. 
Each job has a known processing time and an uncertain job type.
Job types are categorical variables associated with a weight, representing the priority of jobs of that type.
Because the type of each job is unknown when the schedule is constructed, they must first be predicted using a classifier based on features of the jobs.
When all types are known, an optimal job sequence is obtained using the Weighted Shortest Processing Time (WSPT) first rule, which orders jobs from the smallest to the largest ratio of processing time to weight.

We generated instances with $n \in \{100, 200, 300\}$ jobs.
Processing times are sampled independently from $U(1,50)$, and each job is assigned one of four possible types with equal probability.
The weights of the types are evenly distributed over the interval $[1,H]$, where $H \in \{ 8, 16, 32, 64, 128 \}$.
The parameter $H$ controls the heterogeneity of job priorities: larger values create larger differences between the consequences of correctly and incorrectly predicting a job type.
For each combination of $n$ and $H$, we generated 50 problem instances. 

The off-diagonal misclassification probability was allocated uniformly using $h=1$.
The random sampling estimator uses $B=500$ prediction vectors.
For the first-order approximation, the average single-misclassification regret is estimated using at most $G = 100$ samples.

\begin{figure}[htbp]
    \centering
    \includegraphics[width=1\linewidth]{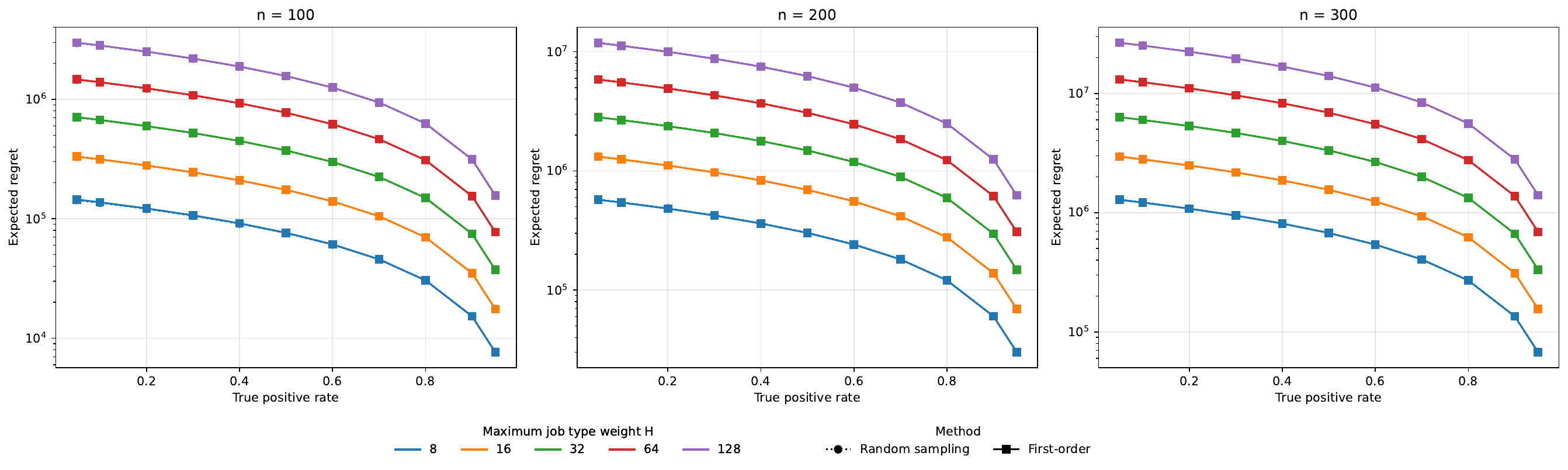}
    \caption{Expected regret as a function of the target TPR, the number of jobs $n$ and the maximum job type weight $H$.}
    \label{fig:scheduling_1}
\end{figure}

Figure \ref{fig:scheduling_1} shows the expected regret as a function of the target TPR for different values of $n$ and $H$.
As expected, regret decreases consistently as predictive accuracy improves.
The magnitude of regret also increases strongly with $H$: when the type weights differ substantially, confusing a high-priority job with a low-priority job can have a much larger effect on the resulting sequence.
Regret furthermore increases with the number of jobs, as larger instances contain more opportunities for misclassification and for the resulting sequencing errors to accumulate.
For all considered settings, the line of the first-order approximation method closely follows the random sampling approach, indicating that the error-to-regret mapping can be estimated accurately without repeatedly sampling complete prediction vectors.

\begin{figure}[htbp]
    \centering
    \includegraphics[width=1\linewidth]{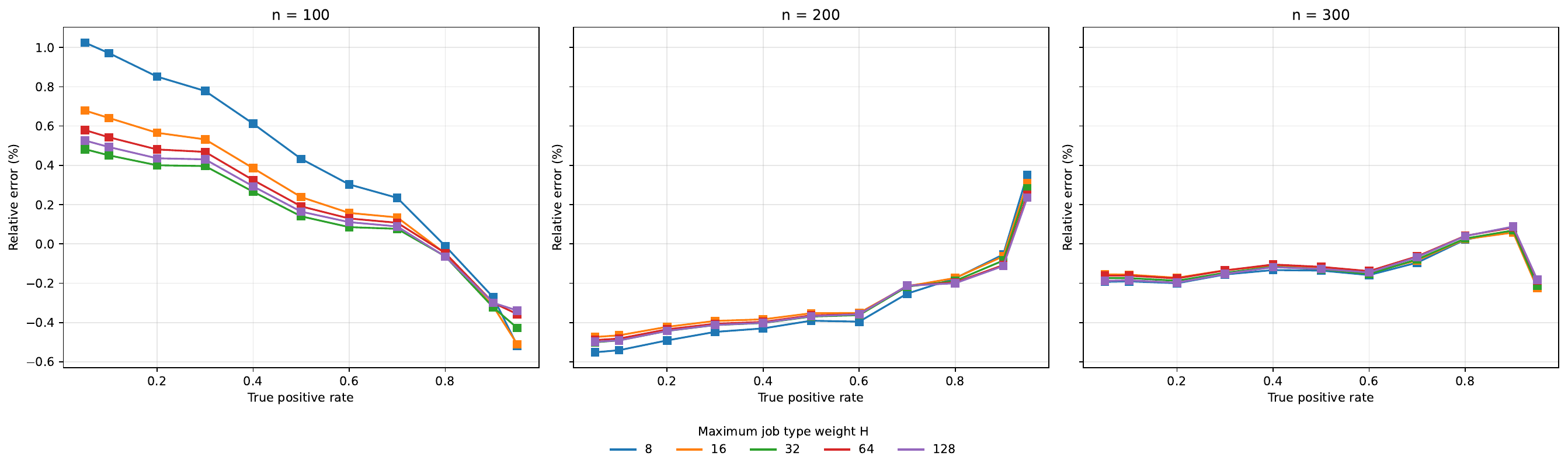}
    \caption{Relative error of the first-order approximation as a function of the target TPR, the number of jobs $n$ and the maximum job type weight $H$.}
    \label{fig:scheduling_2}
\end{figure}

Figure \ref{fig:scheduling_2} analyzes the accuracy of the first-order approximation in more detail by plotting the relative error with respect to the random sampling estimator.
The results confirm that the first-order approximation closely reproduces the random sampling results, with absolute relative errors typically below 0.5\% and substantially smaller errors for $n=300$. 
For $n=100$ the first-order approximation overestimates the random sampling mean for TPR$\leq0.7$, after which it starts to underestimate.  
For $n=200$, the opposite trend is observed: the approximation underestimates the random sampling mean for most of the considered TPR values, approaches zero bias as the TPR increases, and overestimates slightly at 0.95.
For $n=300$, the errors remain closer to zero across the full range of TPR values.
The maximum job type weight has no clear consistent effect on the error.
Moreover, its impact decreases as $n$ increases, as reflected by the near-overlap of the corresponding lines for $n=300$.


\subsubsection{Knapsack with type-dependent item profit}

The second problem is the one-dimensional knapsack problem where a subset of items must be selected to maximize total profit without exceeding a fixed capacity.
Each item has a known weight and belongs to one of several types, with the type determining its profit.
The type of each item is unknown at the time of solving and must therefore be predicted beforehand.
Once the item types are specified, the resulting deterministic knapsack problem is solved using integer programming.

We generated instances containing 100, 200 or 300 items.
The weight of each item was sampled independently from $U(1,20)$, and the knapsack capacity was set to 35\% of the total item weight.
Each item is assigned one of four types with equal probability.
The profits associated with these types are distributed evenly over the interval $[1,H]$, where $H \in \{ 8, 16, 32, 64, 128 \}$.
Similar to the scheduling problem, $H$ controls the heterogeneity of the type-dependent profits: larger values increase the difference between the consequences of misclassifications. 
For each combination of the number of items $n$ and $H$, we generated 50 problem instances.
When generating compatible FPR target values, the off-diagonal misclassification probability was again allocated uniformly.
The same algorithm parameters were used as for the scheduling case: $B=500$ and $G = 100$.

\begin{figure}[htbp]
    \centering
    \includegraphics[width=\linewidth]{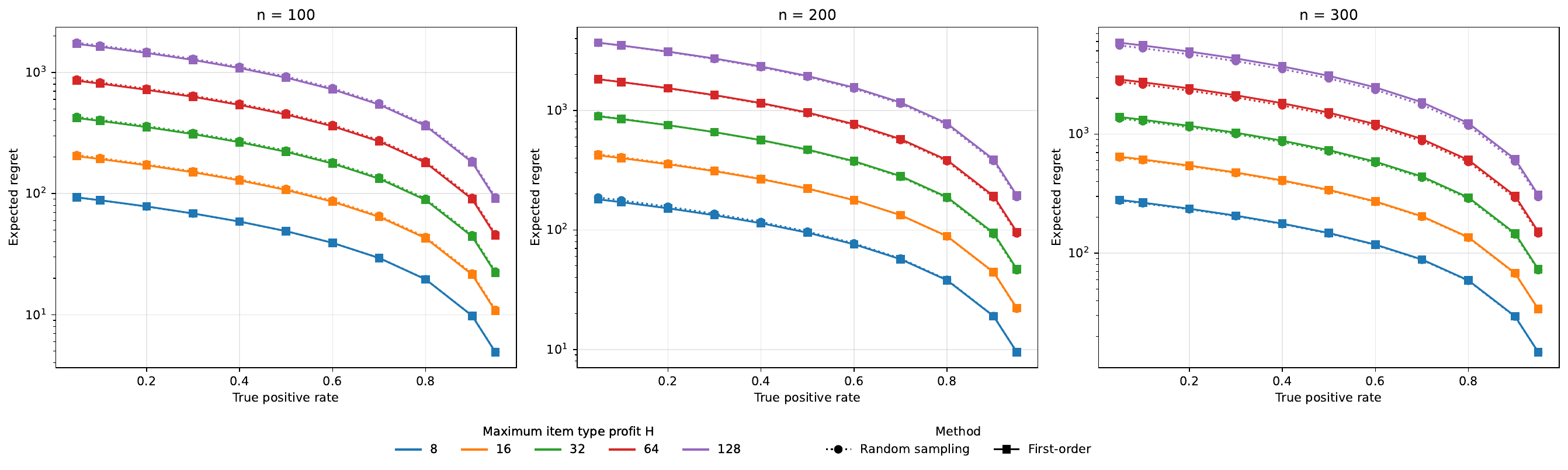}
    \caption{Expected regret as a function of the target TPR, the number of items $n$ and the maximum item type profit $H$.}
    \label{fig:knapsack_1}
\end{figure}

Figure \ref{fig:knapsack_1} shows the expected regret as a function of the target TPR for different values of $n$ and $H$.
In general, the results are similar to those of the scheduling problem.
The expected regret decreases consistently as the target TPR increases, but increases with $H$, particularly at low and intermediate TPR values, where the impact of misclassifying an item becomes larger when the differences between type-dependent profits are larger. 
Larger instances generally produce greater regret because they contain more opportunities for classification errors to influence the selected set.
The first-order approximation follows the random sampling benchmark closely.

\begin{figure}[htbp]
    \centering
    \includegraphics[width=\linewidth]{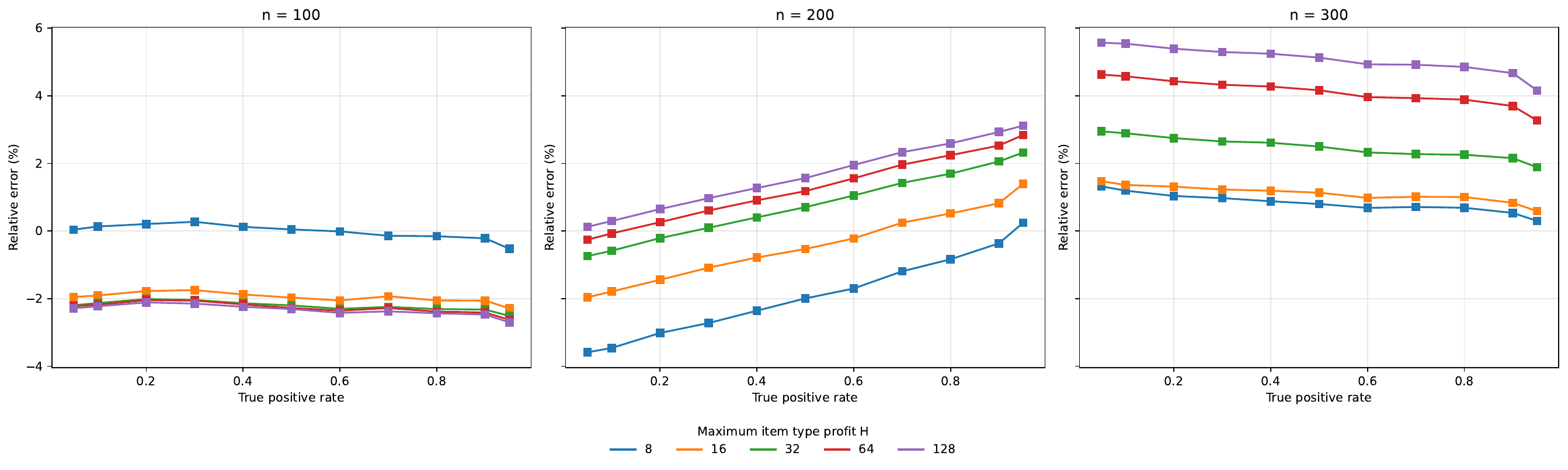}
    \caption{Relative error of the first-order approximation as a function of the target TPR, the number of items $n$ and the maximum item type profit $H$.}
    \label{fig:knapsack_2}
\end{figure}

Figure \ref{fig:knapsack_2} plots the relative error of the first-order approximation with respect to the random sampling estimator for different values of $n$ and $H$.
The first-order approximation generally follows the random sampling estimates closely, although the direction of the approximation bias depends on the parameter setting.
For $n=100$, the approximation is almost exact for a maximum type profit of $8$, while it generally underestimates the random sampling values for larger maximum type profits.
For $n=200$, the bias depends more strongly on the maximum type profit: the approximation tends to underestimate for smaller profits, whereas for larger profits it overestimates as the TPR increases.
For $n=300$, the first-order approximation consistently overestimates the random sampling estimator, with the difference becoming more pronounced as the maximum type profit increases, while generally decreasing as the TPR approaches one.
Overall, the first-order approximation manages to capture the qualitative dependence of the expected regret on the TPR and maximum type profit well.
However, its bias exhibits a clear interaction with both the problem size and the maximum type profit, so that it cannot be treated as a method that uniformly under- or overestimates.

\subsubsection{Shortest path with class-dependent travel time multipliers}

The final problem is a shortest path problem on an urban road network.
Each road has a known base travel time and belongs to one of four road classes.
Each class represents different levels of congestion or delay and is associated with a multiplier that increases the base travel time.
The road classes are the unknown parameter in this problem, and must be predicted before the shortest path can be computed.
Once the road classes are predicted and the updated travel times are known, the shortest path is computed using Dijkstra's algorithm.

\begin{figure}[htbp]
    \centering
    \includegraphics[width=0.6\linewidth]{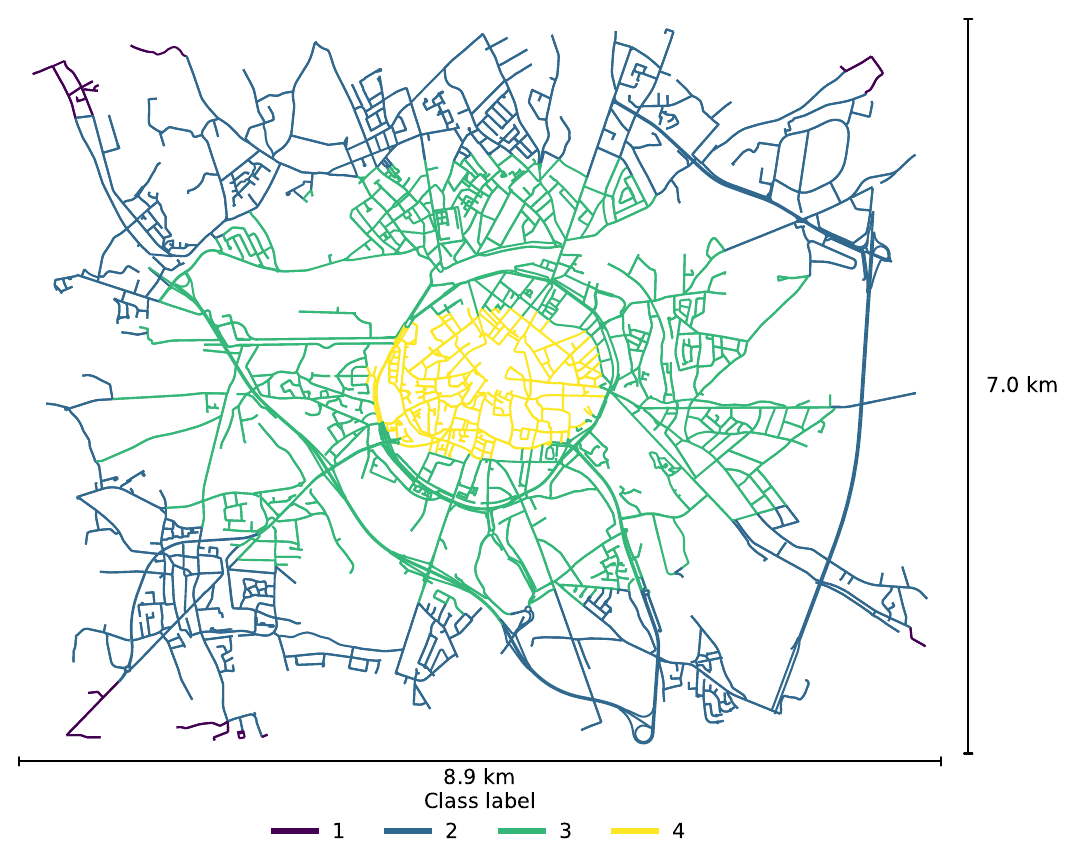}
    \caption{Graph representing the considered road network.}
    \label{fig:sh_graph}
\end{figure}

Using data from OpenStreetMap, we constructed a graph representing the road network in an urban area and its surroundings, covering approximately 62.6km$^2$.
The resulting graph, shown in Figure \ref{fig:sh_graph}, consists of 2626 nodes and 5773 edges.
The true class of each edge is determined by its distance from the center of the considered geographic area.
Edges closer to the center are assigned classes with high multipliers, while edges further away are associated with smaller multipliers.
The multipliers associated with the considered classes are distributed evenly over the interval $[1,H]$, with $H \in \{ 4, 8, 6 \}$.
Increasing $H$ increases heterogeneity in travel times across road classes and therefore also the potential impact of misclassifying an edge.
To generate the problem instances, 50 origin-destination pairs were randomly sampled from the network.
Again, the off-diagonal misclassification probability was allocated uniformly.
The same algorithm parameters as in the previous two experiments were used: $B=500$ and $G = 100$.

\begin{figure}[htbp]
    \centering
    \includegraphics[width=\linewidth]{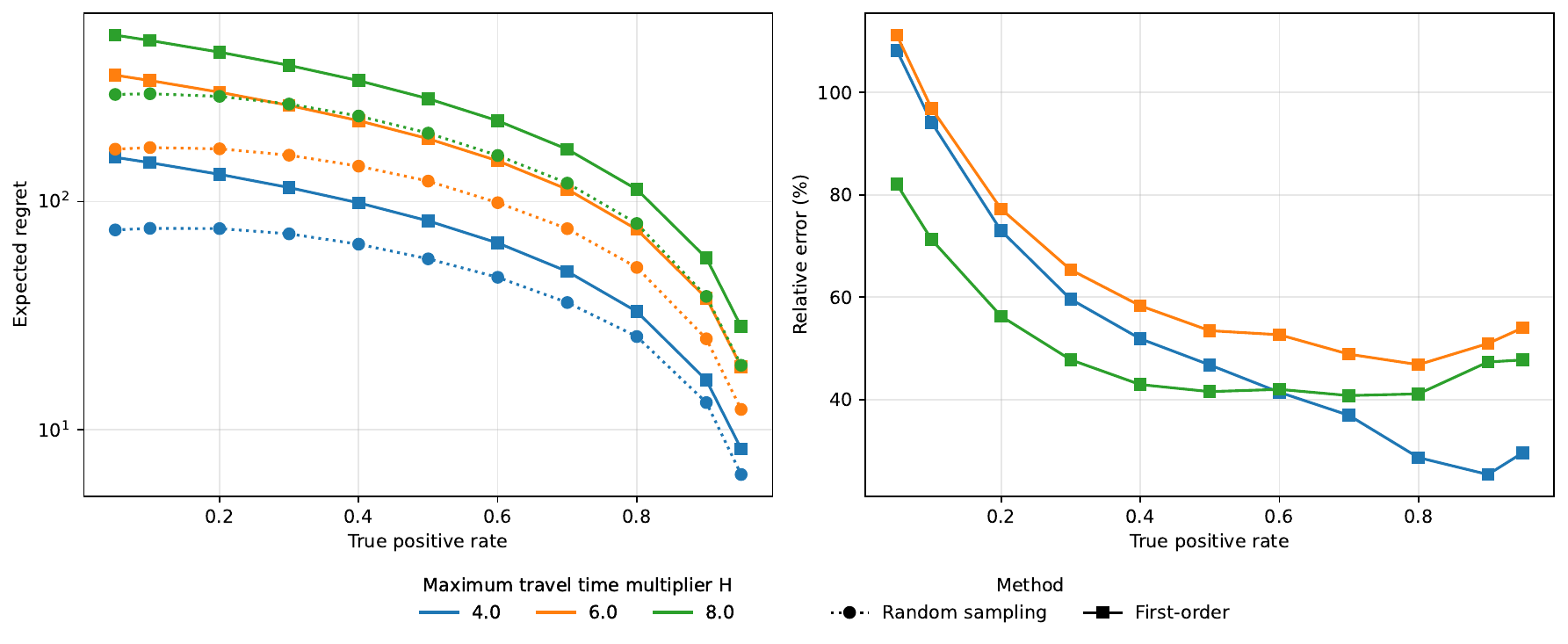}
    \caption{Expected regret and relative error as a function of the target TPR and the maximum travel time multiplier $H$.}
    \label{fig:shortestpath_1}
\end{figure}

Figure \ref{fig:shortestpath_1} compares the expected regret obtained from the random sampling estimator with that of the first-order approximation for different TPR values and maximum travel time multipliers. 
Both methods exhibit the same overall behavior: the expected regret increases with the maximum multiplier and decreases as the TPR becomes large. 
However, in contrast to the previous problems, the first-order approximation consistently overestimates the random sampling estimator across all considered parameter settings. 
The difference is particularly pronounced at low TPR values, where the relative error can exceed 100\%, and generally decreases as the TPR increases. 
The magnitude of the bias also depends on the maximum multiplier, however, without a consistent monotonic relationship. 
These results show that, while the first-order approximation captures the qualitative dependence of the expected regret on the problem parameters, it is substantially less reliable as a quantitative estimator in this setting. 

\subsubsection{Discussion}

The large differences in accuracy of the first-order approximation across the three optimization problems can be related to its main underlying assumption that multiple simultaneous misclassifications can be represented by adding their individual effects.
As discussed in Section \ref{sec:delta_costs}, this assumption becomes less appropriate when the optimization problem has strong combinatorial interactions between errors.
The observed change from very small errors for scheduling, to moderate errors for knapsack and very large errors for shortest path can be interpreted as reflecting the increasing importance of such interactions.

For the single-machine scheduling problem, the consequences of individual prediction errors appear to be relatively close to additive. 
In this setting, a misclassification changes a job's completion time weight and therefore its relative position in the WSPT order. 
The resulting regret is largely driven by the pairwise inversions introduced between that job and the jobs it crosses in the sequence. 
When multiple jobs are misclassified, these effects can often be combined with limited additional interaction, which helps explain why the first-order approximation performs well for the considered problem instances.

The knapsack problem introduces stronger interactions as the inclusion of one item is coupled to the inclusion of all other items through the capacity constraint. 
Changing the predicted profit of one item may therefore cause substitutions between selected and unselected items, and the effect of one error can depend on the profits predicted for other items. 
The sensitivity of the optimal knapsack solution to perturbations in individual profits and weights has been studied by \citet{hifi2005sensitivity}. 
These interactions are consistent with the larger first-order approximation errors. 

Interactions between misclassifications are considerably stronger for the shortest path problem. 
Here, all predicted edge costs jointly determine the path and a change in one or multiple edge costs can cause the optimal solution to switch from one path to an entirely different path. 
Consequently, the effects of individual edge misclassifications may not accumulate independently. 
Two errors that each result in the same path change when considered separately can be redundant when they occur together, while other combinations of errors can reinforce each other. 
The first-order approximation does not account for these effects and instead adds the regret associated with each error separately. 
The effect is particularly pronounced at low TPR values, where many edges are misclassified simultaneously and higher-order interactions are consequently more likely. 
As the TPR increases, fewer simultaneous errors occur and the first-order approximation accuracy improves.


\section{Conclusions and future research}\label{sec:conclusions}

In this paper, we studied ex-ante evaluation of PTO methods when the downstream optimization problem has uncertain categorical parameters that are predicted using multiclass classification.
We introduced an algorithm for simulating multiclass classification predictions with prescribed class-wise TPRs and one-vs-rest FPRs.
We showed that the resulting empirical TPR and FPR values are unbiased for their respective targets and concentrate around these targets as the number of observations increases.
Using this simulation algorithm, we considered two approaches for computing a mapping from classification performance to the expected regret of the downstream optimization problem.
The random sampling approach proposed by \citet{doneda2026robust} provides an unbiased estimate by repeatedly generating complete prediction vectors and solving the corresponding optimization problems. 
However, as this procedure requires solving the optimization problem several times for each TPR-FPR combination included in the mapping, we also proposed a computationally less demanding first-order approximation that estimates expected regret based on the effects of individual misclassifications.

An important practical use of the proposed framework is to translate requirements on downstream decision quality into requirements on predictive performance.
Once the error-to-regret mapping $\Phi$ has been characterized, a regret tolerance $\tau$ can be used to define a required-performance region $\mathcal{A}_{\tau} = \{ (\alpha, \beta) : \Phi(\alpha, \beta; \xi) \leq \tau \}$.
Rather than specifying prediction accuracy requirements independently of their operational consequences, such a region identifies combinations of TPR and FPR values that are sufficient to achieve a desired level of downstream performance. 
This provides an ex-ante criterion for deciding whether the development of a prediction model is worthwhile and, if so, what levels of predictive performance the model should aim to achieve.

In a computational study, we confirmed the finite-sample properties of the proposed classification simulator, thereby demonstrating that the proposed simulation algorithm provides a suitable mechanism for generating predictions at prescribed levels of classification performance.
The two methods for computing the prediction-error-to-regret mapping were evaluated on three downstream optimization problems.
The computational results showed that improving the TPR consistently reduced expected regret.
However, the magnitude and shape of this relationship depended strongly on the optimization problem and its parameters.

The computational study also revealed an important limitation of the first-order approximation.
Although it captured the qualitative relationship between predictive performance and expected regret, its quantitative accuracy varied considerably between problems.
The results indicate that an approximation based only on isolated misclassifications can become inaccurate, depending on the considered optimization problem.
A relevant extension to this approach could therefore be to incorporate pairwise or higher-order interaction effects, ideally only for class pairs or problem elements for which such interactions are most consequential. 
This could retain the computational advantage of the first-order method while improving its accuracy for problems with stronger combinatorial dependencies.
A complementary stream of research could focus on characterizing, theoretically or empirically, conditions under which the first-order approximation is sufficiently accurate.

\section*{Acknowledgments}

This research was supported by KU Leuven C24E/23/012 `Human-centred decision support based on new theory for personnel rostering'.

\section*{CRediT authorship contribution statement}

\textbf{Pieter Smet:} Conceptualization, Formal analysis, Methodology, Software, Visualization, Writing – original draft.

\section*{Declaration of generative AI use}

During the preparation of this work the author used ChatGPT in order to improve the phrasing and readability of text written by the author and to assist in generating code for producing figures. 
After using this tool, the author reviewed and edited the content as needed and takes full responsibility for the content of the published article.


\bibliographystyle{abbrvnat} 
\bibliography{references.bib}


\newpage
\appendix

\section{FPR as a function of $Q$}
\label{appendix:constraint3}

$\fprM_j$ is the probability that class $j$ is predicted while the true class is not j, or $\mathbb{P}(\hat{Y} = j \mid Y \neq j)$.
By the law of total probability, we can rewrite this as
\[
\mathbb{P}(\hat{Y} = j \mid Y \neq j) = \sum_{\substack{i=1\\ i\neq j}}^{K}\mathbb{P}(\hat{Y} = j \mid Y = i,Y \neq j)\mathbb{P}(Y = i \mid Y \neq j).
\]
But if $i \neq j$, then $Y=i$ already implies $Y\neq j$. 
Therefore,
\[
\mathbb{P}(\hat{Y} = j \mid Y = i,Y \neq j) = \mathbb{P}(\hat{Y} = j \mid Y = i).
\]
By definition of the kernel, $\mathbb{P}(\hat{Y} = j \mid Y = i) = Q(j \mid i)$.
Therefore
\[
\mathbb{P}(\hat{Y} = j \mid Y \neq j) = \sum_{\substack{i=1\\ i\neq j}}^{K}Q(j \mid i)\mathbb{P}(Y = i \mid Y \neq j).
\]
Let
\[
\mathbb{P}(Y = i \mid Y \neq j) = \frac{\mathbb{P}(Y = i \cap Y \neq j)}{\mathbb{P}(Y \neq j)}.
\]
Since $i \neq j$, the event $Y=i$ automatically implies $Y\neq j$, so
\[
\mathbb{P}(Y = i \cap Y \neq j) = \mathbb{P}(Y = i) = \pi_i.
\]
Also,
\[
\mathbb{P}(Y \neq j) = 1 - \mathbb{P}(Y = j) = 1 - \pi_j.
\]
Combining these results in
\[
\mathbb{P}(\hat{Y} = j \mid Y \neq j) = \sum_{\substack{i=1\\ i\neq j}}^{K} \frac{\pi_i}{1-\pi_j} Q(j \mid i).
\]

\section{Simulating binary classification predictions}
\label{appendix:binary}

For binary classification, let $\mathcal{Y} = \{0, 1\}$, where label 1 denotes the positive class and label 0 denotes the negative class.
Let $Y \in \mathcal{Y}$ be the true label of an instance and let $\hat{Y} \in \mathcal{Y}$ be the simulated prediction.
The target true positive rate and false positive rate are defined as $\tprM = \mathbb{P}(\hat{Y}=1\mid Y=1)$ and $\fprM = \mathbb{P}(\hat{Y}=1\mid Y=0)$.

Unlike in the multiclass setting, these two target values uniquely determine the conditional confusion kernel $Q$.
If rows are indexed by the true label and columns are indexed by the predicted label, the kernel is 
\[
    Q =
    \begin{pmatrix}
    Q(0 \mid 0) & Q(1 \mid 0) \\
    Q(0 \mid 1) & Q(1 \mid 1) 
    \end{pmatrix}
    =
    \begin{pmatrix}
    1-\fprM & \fprM \\
    1-\tprM& \tprM
    \end{pmatrix}.
\]

Note that the targets could also be expressed as vectors of length two, following the same approach as described in Section \ref{sec:general_def}.
This would result in $\alpha = (1-\fprM, \tprM)$ and $\beta = (1-\tprM, \fprM)$.

Given $Y$ and target TPR and FPR values, Algorithm \ref{alg:bc} generates a simulated prediction without explicitly constructing $Q$.

\begin{algorithm}[htpb]
    \caption{Generating simulated binary predictions}
    \label{alg:bc}
    \begin{algorithmic}[1]
    \Function{SampleBinaryPredictions}{$y_1,\ldots,y_n,\tprM,\fprM$}
    \Require True labels $y_1,\ldots,y_n \in \mathcal{Y}=\{0, 1\}$
    \Require Target values $\tprM, \fprM \in [0, 1 ]$
    \Ensure Simulated predictions $\hat y_1,\ldots,\hat y_n$    
    \For{$r=1,\ldots,n$}
        \State $p \gets U(0,1)$
        \If{$y_r = 1$}
            \State $\hat{y}_r \gets 1$ \textbf{if} $p \leq \tprM$ \textbf{else} $0$
        \Else
            \State $\hat{y}_r \gets 0$ \textbf{if} $p \leq (1 - \fprM)$ \textbf{else} $1$
        \EndIf
    \EndFor    
    \State \Return $\hat y_1,\ldots,\hat y_n$
    \EndFunction
    \end{algorithmic}
\end{algorithm}

\setcounter{theorem}{0}

\section{Proof of Theorem \ref{th:exp_tpr}}\label{app_proof_exp_tpr}

\begin{theorem}[Unbiasedness of the simulated TPR]
    For any class $i$ with $n_i>0$, the predictions generated by Algorithm \ref{alg:mc_p2} satisfy
    \[
    \mathbb{E}[\widehat{\tprM}_i] = \alpha_i.
    \]
\end{theorem}
\begin{proof}
    For each instance $r$ with true label $y_r=i$, we define
    \[
    Z^{(i)}_r = \mathbf{1}\{ \hat{Y}_r = i\}.
    \]
    Since Algorithm \ref{alg:mc_p2} samples predictions from $Q(\cdot\mid i)$, we have
    \[
    \mathbb{P}(Z^{(i)}_r = 1) = \mathbb{P}(\hat{Y}_r = i \mid Y_r = i) = Q(i \mid i).
    \]
    By construction of the kernel, $Q(i \mid i) = \alpha_i$.
    Therefore,
    \[
    Z^{(i)}_r \sim Bernoulli(\alpha_i).
    \]
    The expectation of a random Bernoulli variable is the average of its distribution. Hence,
    \[
    \mathbb{E}[Z^{(i)}_r] = \alpha_i.
    \]
    By rewriting $\widehat{\tprM}_i$ using the Bernoulli variables, we get
    \[
    \widehat{\tprM}_i = \frac{1}{n_i}\sum\limits_{r:y_r=i}Z^{(i)}_r.
    \]
    Now consider the expectation of $\widehat{\tprM}_i$, using linearity of expectation and substituting the expectation of $Z^{(i)}_r$, we get
    \[
    \mathbb{E}[\widehat{\tprM}_i] = \mathbb{E} \left[ \frac{1}{n_i}\sum\limits_{r:y_r=i}Z^{(i)}_r  \right] = \frac{1}{n_i}\sum\limits_{r:y_r=i} \mathbb{E}[Z^{(i)}_r] = \frac{1}{n_i}\sum\limits_{r:y_r=i} \alpha_i.
    \]
    Since there are $n_i$ terms in the sum, each with expectation $\alpha_i$, this can be rewritten as
    \[
    \mathbb{E}[\widehat{\tprM}_i] = \frac{1}{n_i} n_i \alpha_i = \alpha_i.
    \]    
\end{proof}

\section{Proof of Theorem \ref{th:exp_fpr}}\label{app_proof_exp_fpr}

\begin{theorem}[Unbiasedness of the simulated one-vs-rest FPR]
    For any class $j$ with $n_j<n$, the predictions generated by Algorithm \ref{alg:mc_p2} satisfy
    \[
    \mathbb{E}[\widehat{\fprM}_j] = \beta_j
    \]
\end{theorem}

\begin{proof}
    For each instance $r$ with $y_r\neq j$, we define
    \[
    W^{(j)}_r = \mathbf{1} \{ \hat{Y}_r = j\}.
    \]
    If $y_r = i$, where $i \neq j$, then
    \[
    \mathbb{P}(W^{(j)}_r = 1) = \mathbb{P}(\hat{Y}_r = j \mid Y_r = i) = Q(j \mid i).
    \]
    Therefore,
    \[
    W^{(j)}_r \sim Bernoulli(Q(j \mid i)), \qquad \mathbb{E}[W^{(j)}_r] = Q(j \mid i).
    \]
    The variables $W^{(j)}_r$ are not necessarily identically distributed, because $Q(j \mid i)$ may depend on the true class $i$. 
    However, linearity of expectation does not require identical distributions. 
    Therefore,
    \[
    \mathbb{E}[\widehat{\fprM}_j] = \mathbb{E}\left[ \frac{1}{n - n_j} \sum\limits_{r:y_r\neq j}W^{(j)}_r \right] = \frac{1}{n - n_j} \sum\limits_{r:y_r\neq j} \mathbb{E}[W^{(j)}_r].
    \]
    Grouping the sum by true class gives $n_i$ observations with true label $i$, so
    \[
    \mathbb{E}[\widehat{\fprM}_j] =  \frac{1}{n - n_j} \sum\limits_{i \neq j} n_i Q(j \mid i).
    \]
    Using $Q(j \mid i) = M_{ij}/\pi_i$ and $\pi_i = n_i / n$, we obtain
    \[
    n_i Q(j \mid i) = n_i \frac{M_{ij}}{\pi_i}  = n_i \frac{M_{ij}}{n_i / n} = n M_{ij}.
    \]
    Therefore,
    \[
    \sum\limits_{i \neq j} n_i Q(j \mid i) = \sum\limits_{i \neq j} n M_{ij} =  n \sum\limits_{i \neq j}  M_{ij}.
    \]

    By construction of the off-diagonal joint mass matrix $M$, the column-demand constraint enforces
    \[
    \sum\limits_{i \neq j}M_{ij} = d_j = (1 - \pi_j)\beta_j.
    \]
    It follows that
    \[
    \sum\limits_{i \neq j}n_iQ(j \mid i) = n(1 - \pi_j)\beta_j.
    \]
    Substituting this into the expectation results in
    \[
    \mathbb{E}[\widehat{\fprM}_j] =  \frac{1}{n - n_j}n(1 - \pi_j)\beta_j.
    \]
    Since $\pi_j = n_j / n$, we have $n(1 - \pi_j) = n - n_j$.
    So we can simplify the expectation to
    \[
    \mathbb{E}[\widehat{\fprM}_j] = \beta_j
    \]    
\end{proof}

\section{Proof of Theorem \ref{th:conc_tpr}}\label{app_proof_conc_tpr}

\begin{theorem}[Concentration of the simulated TPR]
    For any class $i \in \mathcal{Y}$ with $n_i>0$ and any $\epsilon > 0$, the predictions generated by Algorithm \ref{alg:mc_p2} satisfy
    \[
    \mathbb{P}\left( |\widehat{\tprM}_i - \alpha_i| \geq \epsilon \right) \leq 2 \exp(-2n_i\epsilon^2)
    \]
\end{theorem}

\begin{proof}
    For every observation $r$ whose true label is $y_r=i$, define the random variable $Z^{(i)}_r = \mathbf{1}\{ \hat{Y}_r = i \}$.
    From the proof of Theorem \ref{th:exp_tpr}, we know that the empirical TPR can be rewritten as an average of independent bounded random variables $Z^{(i)}_r$
    \[
     \widehat{\tprM}_i = \frac{1}{n_i} \sum\limits_{r:y_r=i} Z^{(i)}_r.
    \]    
    From Theorem \ref{th:exp_tpr}, we know that $\mathbb{E}[\widehat{\tprM}_i] = \alpha_i$.
    So we can write
    \[
    \widehat{\tprM}_i - \alpha_i = \widehat{\tprM}_i - \mathbb{E}[\widehat{\tprM}_i].
    \]

    We now apply Hoeffding’s inequality \citep{hoeffding1963probability} to the average of $n_i$ independent random variables bounded in $[0,1]$. 
    For the upper tail,
    \[
    \mathbb{P}(\widehat{\tprM}_i - \mathbb{E}[\widehat{\tprM}_i] \geq \epsilon) \leq \exp(-2n_i\epsilon^2).
    \]
    Since $\mathbb{E}[\widehat{\tprM}_i] = \alpha_i$, this is
    \[
    \mathbb{P}(\widehat{\tprM}_i - \alpha_i \geq \epsilon) \leq \exp(-2n_i\epsilon^2).
    \]

    Similarly, Hoeffding’s inequality gives the lower-tail bound
    \[
    \mathbb{P}(\widehat{\tprM}_i - \alpha_i \leq -\epsilon) \leq \exp(-2n_i\epsilon^2).
    \]

    Now observe that $|\widehat{\tprM}_i - \alpha_i| \geq \epsilon$ is equivalent to the union of two events $\widehat{\tprM}_i - \alpha_i \geq \epsilon$ and $\widehat{\tprM}_i - \alpha_i \leq -\epsilon$.
    Therefore, by the union bound,
    \begin{align*}
       \mathbb{P}(|\widehat{\tprM}_i - \alpha_i| \geq \epsilon)  & \leq   \mathbb{P}(\widehat{\tprM}_i - \alpha_i \geq \epsilon) + \mathbb{P}(\widehat{\tprM}_i - \alpha_i \leq -\epsilon) \\
                                                                 & \leq \exp(-2n_i\epsilon^2) + \exp(-2n_i\epsilon^2) \\
                                                                 & \leq 2 \exp(-2n_i\epsilon^2)
    \end{align*}
\end{proof}

\section{Proof of Theorem \ref{th:conc_fpr}}\label{app_proof_conc_fpr}

\begin{theorem}[Concentration of the simulated FPR]
    For any class $j \in \mathcal{Y}$ with $n_j < n$ and any $\epsilon > 0$, the predictions generated by Algorithm \ref{alg:mc_p2} satisfy
    \[
    \mathbb{P}\left( |\widehat{\fprM}_j - \beta_j| \geq \epsilon \right) \leq 2 \exp(-2(n-n_j)\epsilon^2)
    \]
\end{theorem}

\begin{proof}
    For each observation $r$ with $y_r\neq j$, define the random variable $W_r^{(j)}=\mathbf{1}\{\hat Y_r=j\}$.
    If $y_r=i\neq j$, then
    \[
    W_r^{(j)}\sim Bernoulli(Q(j\mid i)).
    \]
    The variables $\{W_r^{(j)}:y_r\neq j\}$ are independent because Algorithm \ref{alg:mc_p2} samples predictions independently across observations. 
    They are not necessarily identically distributed, since $Q(j\mid i)$ may depend on the true class $i$, but each variable is bounded in $[0,1]$. 
    Moreover,
    \[
    \widehat{\fprM}_j = \frac{1}{n-n_j}\sum\limits_{r:y_r\neq j}W^{(j)}_r.
    \]
    Using Theorem \ref{th:exp_fpr} and applying Hoeffding’s inequality similar to the proof of Theorem \ref{th:conc_tpr}, but now to the average of $n-n_j$ independent random variables bounded in $[0,1]$, the upper-tail bound is
    \[
    \mathbb{P}\left( \widehat{\fprM}_j - \beta_j \geq \epsilon \right) \leq \exp(-2(n-n_j)\epsilon^2).
    \]
    Similarly, Hoeffding’s inequality gives the lower-tail bound
    \[
     \mathbb{P}\left( \widehat{\fprM}_j - \beta_j \leq -\epsilon \right) \leq \exp(-2(n-n_j)\epsilon^2).
    \]
    Using the union bound,
    \begin{align*}
     \mathbb{P}\left( |\widehat{\fprM}_j - \beta_j| \geq \epsilon \right) &  \leq \mathbb{P}(\widehat{\fprM}_j - \beta_j \geq \epsilon) + \mathbb{P}(\widehat{\fprM}_j - \beta_j \leq -\epsilon) \\     
     &\leq \exp(-2(n-n_j)\epsilon^2) + \exp(-2(n-n_j)\epsilon^2) \\
     &\leq 2 \exp(-2(n-n_j)\epsilon^2)
    \end{align*}
\end{proof}

\end{document}